\documentclass[conference,compsoc]{IEEEtran}
\usepackage[T1]{fontenc}
\usepackage[utf8]{inputenc}
\usepackage{amsmath,amssymb}
\usepackage{booktabs,tabularx,array,multirow}
\usepackage{graphicx}
\usepackage{placeins}
\graphicspath{{figures/}{figures/plots/}}
\usepackage{tikz}
\usetikzlibrary{arrows.meta,positioning,fit}
\usepackage{xurl}
\usepackage[nocompress]{cite}
\usepackage[hidelinks]{hyperref}

\title{When Valid Tool Calls Change Meaning:\\Formation-Consistent Dispatch for LLM Agents}
\author{
  \IEEEauthorblockN{Geonwoo Kim and Brent ByungHoon Kang}
  \IEEEauthorblockA{Korea Advanced Institute of Science and Technology (KAIST)\\
  \texttt{signal@kaist.ac.kr}}
}
\hypersetup{
  pdftitle={When Valid Tool Calls Change Meaning: Formation-Consistent Dispatch for LLM Agents},
  pdfauthor={Geonwoo Kim and Brent ByungHoon Kang}
}

\begin{document}
\maketitle
\begin{abstract}
Tool-enabled agents form calls from model-visible interfaces, while hosts later select their
implementation. Standard dispatch omits the descriptor--handler
relation. An unchanged and schema-valid call can therefore acquire a different security effect
during rollout, reconnect, or delayed approval. We call this failure \emph{schema-epoch drift}.

We present \emph{formation-consistent dispatch} (FCD), which connects implementation analysis to
execution authority. Reviewed profiles produce provenance-bound over-approximations of declared
in-scope effects from official source. Under a closed-target approval policy, a verifier applies
each formed call to a summary and captures a successor only when its effects fit the call's security
contract. Atomic admission and a final-hop fence preserve this decision to the effect. The exact
source retains priority. The captured successor becomes eligible only after source retirement.

Stock releases and deployment changes reproduced the failure. Four profiles covered 32 official
releases: 29 required no release-specific change and three escalated. A frozen 16-release expansion
matched a separate source oracle. In a preregistered stock comparison, FCD completed all three
pending calls whose effect remained private and blocked all three whose omission became public.
Exact pinning and release-wide denial stopped all six calls. Release-wide approval completed all six
but produced three public effects. A separate lifecycle experiment carried a formation-captured
certificate across source retirement. The same safe certificate installed later governed new
formations without expanding the pending call's authority.
\end{abstract}

\section{Introduction}
\label{sec:introduction}

Tool-enabled agents separate call formation from execution. An MCP host gives the model a tool
name, description, and input schema. The model returns a name and arguments
~\cite{mcp-tools,mcp-ts-client}. The host later selects the binary, configuration, and replica that
handle the call. Approval, retry, reconnect, and rollout can occur between these two decisions.

The ordinary \texttt{tools/call} payload carries the name and arguments. It does not identify the
descriptor or handler that gave those bytes their approved meaning. Protocol negotiation identifies
an MCP communication revision rather than a tool implementation
~\cite{mcp-basic,mcp-versioning}. The model therefore cannot verify that execution uses the
implementation whose interface shaped its call.

This gap allows an unchanged and schema-valid call to acquire a different security effect. A model
may see a descriptor in which omitting a field requests a private resource. It can correctly omit
that field. A rollout can then route the retained call to a handler that interprets omission as
public. The model and any argument policy still observe a valid call, while the selected interpreter
reverses its effect.

We call this temporal-integrity failure \emph{schema-epoch drift}. A schema epoch binds a
model-visible descriptor to an implementation and its security-relevant execution context. Drift
occurs when a call formed under one epoch executes under another that assigns the same bytes a
different security meaning. The descriptor itself may remain unchanged when the handler or its
execution context changes. Ordinary operations can trigger the failure. A controller with rollout
or routing authority can also induce it by changing availability or reconnect timing after
formation.

Existing agent defenses validate definitions, arguments, servers, policies, and execution
capsules~\cite{parasites-mcp,agent-permissions,etdi,attested-admission,tool-forge}. ETDI also stores
the approved definition version and hash. A changed or new definition requires
reapproval~\cite{etdi}. This preserves approval--definition integrity. FCD extends the protected
relation to the declared in-scope effects that a concrete successor assigns to one pending call. It also
fixes when that successor may inherit the call after source retirement. The closest agent TOCTOU work
studies changes between separate model-issued calls~\cite{mind-the-gap}. Schema-epoch drift changes
the interpreter inside one formation-to-execution interval, after the model has fixed the call
bytes.

Version pinning records which implementation a pending call names. Draining and fencing preserve
lifecycle order and selected identity. Authorized substitution requires three further decisions.
A profile determines whether the target preserves the concrete call's effect contract. The
formation record determines whether this invocation authorized that target. Lifecycle state
determines when the captured target becomes eligible to execute. A successor can be
contract-compatible without being authorized for this invocation. It can be authorized without yet
being eligible to execute. FCD preserves all three decisions from formation to effect.

This distinction applies when approval covers interpreter identity as well as permitted effects. A
deployment that delegates at formation to all future contract-compatible implementations adopts a
different authorization policy.

Semantic differencing and change-impact analysis identify behavioral changes between program
versions~\cite{symdiff,idise}. They do not turn those findings into invocation authority or preserve
the decision through deployment. FCD connects a scoped effect analysis to the execution path of one
formed call.

We introduce \emph{formation-consistent dispatch} (FCD). A reviewed profile extracts
security-relevant facts from official implementation source. It converts those facts into a
provenance-bound summary within a declared source, argument, and effect envelope. At formation, a verifier substitutes the concrete arguments and
authenticated context. It captures a successor only when every summarized effect fits the call's
security contract. This decision distinguishes safe calls from unsafe calls within the same
release transition.

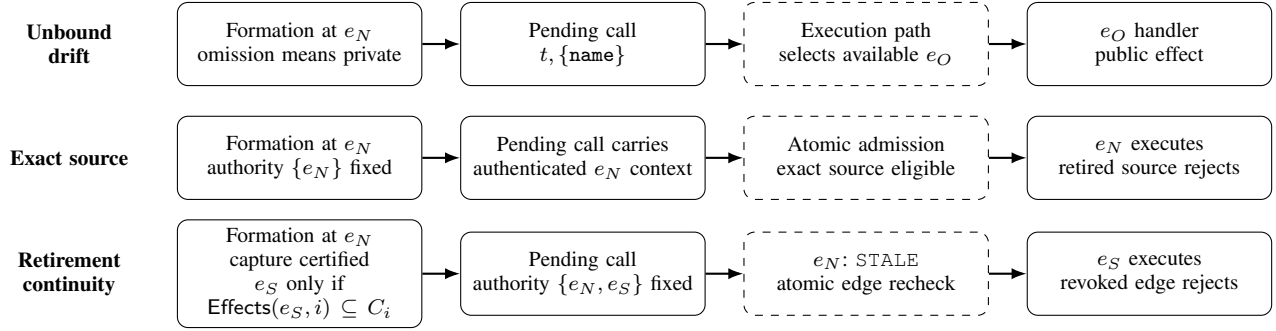
\begin{figure*}[t]
\centering
\begin{tikzpicture}[
  node distance=7mm and 5mm,
  every node/.style={font=\footnotesize,align=center},
  stage/.style={draw,rounded corners,text width=0.17\textwidth,minimum height=11mm,inner sep=3pt},
  decision/.style={draw,dashed,rounded corners,text width=0.17\textwidth,minimum height=11mm,inner sep=3pt},
  flow/.style={-{Latex[length=2mm]},thick},
  label/.style={font=\footnotesize\bfseries,anchor=east,text width=0.09\textwidth}
]
\node[label] (unsafe-label) {Unbound drift};
\node[stage,right=of unsafe-label] (uform) {Formation at $e_N$\\omission means private};
\node[stage,right=of uform] (ucall) {Pending call\\$t,\{\mathtt{name}\}$};
\node[decision,right=of ucall] (uroute) {Execution path selects available $e_O$};
\node[stage,right=of uroute] (ueffect) {$e_O$ handler\\public effect};
\draw[flow] (uform) -- (ucall);
\draw[flow] (ucall) -- (uroute);
\draw[flow] (uroute) -- (ueffect);

\node[label,below=9mm of unsafe-label] (safe-label) {Exact source};
\node[stage,right=of safe-label] (bform) {Formation at $e_N$\\authority $\{e_N\}$ fixed};
\node[stage,right=of bform] (bcall) {Pending call carries authenticated $e_N$ context};
\node[decision,right=of bcall] (bgate) {Atomic admission\\exact source eligible};
\node[stage,right=of bgate] (beffect) {$e_N$ executes\\retired source rejects};
\draw[flow] (bform) -- (bcall);
\draw[flow] (bcall) -- (bgate);
\draw[flow] (bgate) -- (beffect);

\node[label,below=9mm of safe-label] (cont-label) {Retirement continuity};
\node[stage,right=of cont-label] (cform) {Formation at $e_N$\\capture certified $e_S$ only if\\$\mathsf{Effects}(e_S,i)\subseteq C_i$};
\node[stage,right=of cform] (ccall) {Pending call\\authority $\{e_N,e_S\}$ fixed};
\node[decision,right=of ccall] (cgate) {$e_N$: \texttt{STALE}\\atomic edge recheck};
\node[stage,right=of cgate] (ceffect) {$e_S$ executes\\revoked edge rejects};
\draw[flow] (cform) -- (ccall);
\draw[flow] (ccall) -- (cgate);
\draw[flow] (cgate) -- (ceffect);
\end{tikzpicture}
\caption{Formation-to-execution drift and FCD's two authorized paths. Formation captures a
contract-compatible successor and fixes invocation authority. Atomic admission executes the exact
source while eligible and selects the captured successor only after safe retirement.}
\label{fig:drift-overview}
\end{figure*}

FCD preserves that source-derived decision through the lifecycle. It fixes the exact source and one
authorized successor when the host forms the call. Later decisions may remove an interpreter from
this set, but they cannot add one. The gateway prefers the source while it remains eligible. It may
select the captured successor only after the source completes retirement. Atomic admission orders
the call against drain and revocation. A durable lease pins the selected pool. A final-hop identity
fence verifies the artifact and configuration before the protected effect. The successor may enter
the runtime registry later, but its exact artifact, configuration, summary, and certificate must
already be part of the formation record. A different target requires a new formation and any
configured action approval.

The trusted host derives the call's contract from the concrete action and any external approval.
Each profile states its supported source structure, argument domain, declared in-scope effects, and external
assumptions. Unsupported structures and arguments require exact execution, manual review, or a new
call. The same lifecycle enforces manually reviewed compatibility edges and generated call-specific
certificates.

Our evaluation connects three forms of evidence. Stock releases and deployment paths establish the
failure. Four reviewed profiles and stock-handler traces test call-specific effect evidence. A
preregistered policy comparison shows that FCD completes safe pending calls while blocking calls
whose meaning changed. Lifecycle and replicated-gateway experiments carry that decision from
formation to effect.

This paper makes four contributions:
\begin{enumerate}
  \item It identifies schema-epoch drift and demonstrates the failure across stock releases,
  reconnects, and orchestrated rollback.
  \item Under a closed-target approval policy, it separates call-specific effect compatibility,
  invocation authority, and lifecycle eligibility. FCD authorizes substitution only at their
  intersection and prevents later policy from expanding a pending call's authority.
  \item It derives provenance-bound, profile-bounded effect summaries for concrete calls. It checks recognized callee
  identities against reviewed inventories in bounded handler regions. The resulting certificates
  distinguish safe from unsafe calls within one release transition.
  \item It evaluates the combined design across official releases, stock-handler end-to-end
  traces, lifecycle races, agent frameworks, and replicated deployments.
\end{enumerate}

\section{Background}
\label{sec:model}

\subsection{The agent tool path}

An MCP host discovers tools through \texttt{tools/list} and gives the model each tool's name,
description, and input schema. The model returns a name and arguments through
\texttt{tools/call}~\cite{mcp-tools,mcp-ts-client}. The host separately selects the binary,
configuration, and replica that execute the request. The ordinary call payload does not carry the
descriptor snapshot or handler identity.

MCP negotiation identifies a communication revision rather than a tool release
~\cite{mcp-basic,mcp-versioning}. Approval, retry, reconnect, rollout, or retirement can therefore
change the selected handler after formation. A call may remain schema-valid while a different
handler changes its visibility, destination, access, or mutation effect.

This separation creates schema-epoch drift when execution selects a different descriptor--handler
relation after formation.

\subsection{Failure and adversary model}

We protect a pending invocation's formation provenance and execution target. An unauthorized effect
comes from an interpreter outside the authority recorded at formation. That interpreter may remain
valid for new calls; the violation is its use for an older call.

Both operations and an adversary can produce this schedule. A rollout, rollback, retry, or reconnect
can select the wrong epoch. A controller with deployment or routing authority can induce the same
condition by withdrawing replicas or changing reconnect timing. For example, it can remove GitHub
v1.4 during approval and route its name-only call to authenticated v1.3. The call and approval stay
unchanged, but the older handler changes private intent to public intent.

The adversary controls availability, unauthenticated service membership, route selection, rollout
direction, and reconnect timing among deployed endpoints. It cannot forge host, catalog, workload,
or compatibility credentials. It cannot alter authenticated registry state, signed manifests,
policy-clock decisions, or the effect-owning guard. Complete mediation keeps execution within
registered pools. The selected handler can be an honest release with different semantics.

The trusted authorization plane retains formation state, authenticates registered targets, orders
lifecycle decisions, and connects admission to the final effect. The registrar defines the
artifact, configuration, flags, and dependencies that identify an interpreter. Profile authors
define supported source structures, arguments, effects, and assumptions. An independent verifier
regenerates each summary, while a reviewer handles cases outside a profile. Table~\ref{tab:tcb}
assigns these roles.

\begin{table}[t]
\centering
\footnotesize
\setlength{\tabcolsep}{3pt}
\caption{Trusted authorization plane. The deployment and routing controller remains outside this TCB.}
\label{tab:tcb}
\begin{tabularx}{\columnwidth}{@{}>{\raggedright\arraybackslash}p{0.25\columnwidth}X>{\raggedright\arraybackslash}p{0.22\columnwidth}@{}}
\toprule
Component & Trusted role & Supported claim \\
\midrule
Host, session channel, and continuation store & Retain the formation snapshot, bind the arguments,
and reject stored-record rollback or substitution & Formation provenance \\
Gateway and registry & Validate credentials and order admission against lifecycle changes & Target identity \\
Pool mapping and final fence & Connect the admitted identity to the implementation that produces the effect & Execution identity \\
Policy clock and resolver & Order expiry, revocation, uncertain outcomes, and safe retirement & Lifecycle safety \\
Registrar, profile author, and compatibility issuer & Define interpreter coverage and analysis
scope; establish source fidelity; review escalations & Conditional semantics \\
\bottomrule
\end{tabularx}
\end{table}

Exact-only dispatch omits the compatibility issuer. A single-host deployment uses SQLite and a
fixed pool mapping. A replicated deployment substitutes a linearizable registry and shared
lifecycle authority. Mutable routing also requires an authenticated final guard. Semantic
containment adds the registrar and either a reviewed profile or manual compatibility evidence.

\section{Case Study: Schema-Epoch Drift in MCP}
\label{sec:cases}

This section establishes three properties of schema-epoch drift. Stock releases can assign different
security effects to the same accepted call, even when their descriptors match. A hosted runtime can
retain a formed call across a catalog update. Ordinary routing and rollback can then deliver that call
to a different interpreter. Local recorders capture the resulting request intent or effect without
creating production resources.

\subsection{Stock release semantics}
\label{sec:release-cases}

\paragraph{Descriptor-changing defaults}
GitHub MCP Server v1.3.0 and v1.4.0 both accept a name-only
\texttt{create\_repository} call. Version 1.4 maps an omitted \texttt{private} field to private
creation, while v1.3 maps the same omission to public creation
~\cite{github-default,github-release-old,github-release-new}. Stock handlers and a REST recorder
confirmed both effects. An explicit \texttt{private:true} control preserved private intent across
the versions.

Azure MCP Server \texttt{2.0.0-beta.32} introduced \texttt{public-network-access} with a false
default. Its predecessor, beta.31, emits \texttt{publicNetworkAccess: Enabled} for the same omitted
argument~\cite{azure-default,azure-release-old,azure-release-new}. Thirty paired resource names
confirmed disabled intent on beta.32 and enabled intent when the beta.32 call reached beta.31.

\paragraph{Descriptor-stable handlers}
Reference Git MCP \texttt{2025.11.25} and \texttt{2025.12.18} expose the same relevant descriptor.
The older handler interprets a fixed argument as a Git option and creates a canary file; the patched
handler rejects it in 3/3 paired runs~\cite{reference-git-option-advisory}. DBHub \texttt{0.22.2}
changes database state under read-only mode, while \texttt{0.22.6} rejects the same call with
\texttt{READONLY\_VIOLATION} in 5/5 runs~\cite{dbhub-readonly-advisory}. These cases show that
descriptor comparison and relisting cannot detect handler or policy changes hidden behind an
unchanged schema.

\begin{table*}[t]
\centering\footnotesize
\renewcommand{\arraystretch}{0.85}
\caption{Representative stock-release and reconnect results. O and N denote older and newer
releases. V marks preserved formation authority; X marks a reversal.}
\label{tab:release-cases}
\begin{tabularx}{\textwidth}{@{}>{\raggedright\arraybackslash}p{0.08\textwidth}>{\raggedright\arraybackslash}p{0.14\textwidth}>{\raggedright\arraybackslash}p{0.10\textwidth}>{\raggedright\arraybackslash}p{0.15\textwidth}Xc@{}}
\toprule
System & Formation $\to$ execution & Relevant argument & Formation-time authority & Recorded execution & Preserved? \\
\midrule
GitHub & N $\to$ O & omitted & private & public & \textbf{X} \\
GitHub & N $\to$ O, bound & omitted & private & no repository request & \textbf{V} \\
Azure & N $\to$ O & omitted & network disabled & network enabled & \textbf{X} \\
Azure & N $\to$ O, bound & omitted & network disabled & no ARM request & \textbf{V} \\
Ref. Git & N $\to$ N & identical bytes & reject; no file & rejected; no file, 3/3 & \textbf{V} \\
Ref. Git & N $\to$ O & identical bytes & reject; no file & canary file created, 3/3 & \textbf{X} \\
DBHub & N $\to$ N & identical bytes & read-only; no change & \texttt{READONLY\_VIOLATION}, 5/5 & \textbf{V} \\
DBHub & N $\to$ O & identical bytes & read-only; no change & database changed, 5/5 & \textbf{X} \\
OpenClaw & exact session & no filter & formation session & empty formation session & \textbf{V} \\
OpenClaw & reconnected session & no filter & formation session & other session's seeded task & \textbf{X} \\
\bottomrule
\end{tabularx}
\end{table*}

OpenClaw isolates execution-context rebinding rather than release-induced semantics. Stock versions
1.7.0 and 1.6.0 exposed identical relevant descriptors. Local \texttt{qwen3:8b} formed
\texttt{openclaw\_task\_list \{\}} in an empty 1.7.0 session. Payload-blind reconnect selected a
different session containing one seeded task in 3/3 runs; exact execution returned the empty
formation session in 3/3. Descriptor and argument digests remained identical.

\subsection{Hosted-model update path}
\label{sec:hosted-synthetic}

We used one synthetic tool and two local effect listeners to test the complete model--runtime path.
Tool discovery, hosted-model formation, Codex permission handling, update notification, and
\texttt{tools/call} followed the product path. \texttt{SAFE} denotes the effect authorized by the
formation-time descriptor; \texttt{PROTECTED} denotes the alternate execution-time effect.

Hosted \texttt{gpt-5.6-sol} formed \texttt{\{\}} under a v1 descriptor that mapped an omitted
\texttt{recipient} to \texttt{SAFE}. During permission delay, v2 changed the default to
\texttt{PROTECTED} and emitted a tool-list notification. The runtime retained the formed
arguments. Four update patterns produced the v2 effect in all 12 held calls. Six stable or
compatible controls produced \texttt{SAFE}, while two strict controls stopped without effect.
This experiment establishes end-to-end reachability through a hosted model and runtime.

\subsection{Deployment reachability}
\label{sec:reachability}

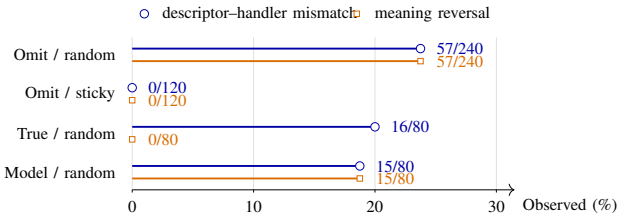
\begin{figure}[t]
\centering
\resizebox{0.97\columnwidth}{!}{%
\begin{tikzpicture}[x=0.18cm,y=0.58cm,font=\scriptsize]
  \foreach \x in {0,10,20,30} {
    \draw[black!12] (\x,-0.45) -- (\x,3.45);
    \draw (\x,-0.45) -- +(0,-0.07) node[below] {\x};
  }
  \draw[->] (0,-0.45) -- (31.5,-0.45) node[below right] {Observed (\%)};
  \node[anchor=east] at (-0.7,3.0) {Omit / random};
  \node[anchor=east] at (-0.7,2.0) {Omit / sticky};
  \node[anchor=east] at (-0.7,1.0) {True / random};
  \node[anchor=east] at (-0.7,0.0) {Model / random};

  \foreach \x/\y/\lab in {23.75/3.16/{57/240},0/2.16/{0/120},20/1.16/{16/80},18.75/0.16/{15/80}} {
    \draw[blue!65!black,thick] (0,\y) -- (\x,\y);
    \node[circle,draw=blue!65!black,fill=white,inner sep=1.25pt] at (\x,\y) {};
    \node[anchor=west,text=blue!65!black] at (\x+0.65,\y) {\lab};
  }
  \foreach \x/\y/\lab in {23.75/2.84/{57/240},0/1.84/{0/120},0/0.84/{0/80},18.75/-0.16/{15/80}} {
    \draw[orange!85!black,thick] (0,\y) -- (\x,\y);
    \node[rectangle,draw=orange!85!black,fill=white,inner sep=1.15pt] at (\x,\y) {};
    \node[anchor=west,text=orange!85!black] at (\x+0.65,\y) {\lab};
  }

  \node[circle,draw=blue!65!black,fill=white,inner sep=1.25pt] at (1.0,4.05) {};
  \node[anchor=west] at (1.8,4.05) {descriptor--handler mismatch};
  \node[rectangle,draw=orange!85!black,fill=white,inner sep=1.15pt] at (18.5,4.05) {};
  \node[anchor=west] at (19.3,4.05) {meaning reversal};
\end{tikzpicture}%
}
\caption{Mixed-version reachability. Omitted calls reverse meaning whenever routing selects the old
handler. Sticky routing prevents mismatch, while an explicit argument prevents reversal.}
\label{fig:reachability}
\end{figure}

Two old and two new stock GitHub replicas ran behind HAProxy's payload-blind
\texttt{random(2)} policy. Each new-descriptor/old-handler match produced public intent. Sticky
routing removed the mismatch, and explicit \texttt{private:true} preserved private intent despite
it. A hosted-model run retained the name-only arguments and reproduced the reversal
(Figure~\ref{fig:reachability}).

A second deployment connected ContextForge v1.0.9's persistent catalog~\cite{contextforge} to a
stable Kubernetes Service and four replicas. The model formed a name-only call from the imported
v1.4 descriptor. A default \texttt{RollingUpdate} replaced all backends with v1.3 before release
~\cite{kubernetes-deployment,kubernetes-service}. Stable controls produced private intent; every
held call produced public intent after rollback. The stable endpoint therefore preserved the tool
name but not the interpreter identity.

\section{Formation-Consistent Tool Dispatch}
\label{sec:defense}

\subsection{Security objective}
\label{sec:invariant}

FCD separates three decisions for a pending call. Effect compatibility determines whether a
successor preserves the call's contract. Invocation authority records whether formation captured
that successor. Lifecycle eligibility permits the successor only after source retirement. A
pending call executes on a successor only at the intersection of these decisions. In the base
deployment, authenticated host policy creates and retains the formation record. A deployment with
an external action-policy layer binds its decision to the same arguments, catalog snapshot, and
interpreter set. The source interpreter receives priority while it remains executable. After safe
retirement, the prototype gateway may select one signed successor captured at formation. Expiry and
revocation before admission remove that option. Once admission commits, its lease wins the registry
order. A deployment that applies later revocation carries the checked generation to an
effect-owning final guard. A successor edge published after formation governs new records.

Compatibility certification and invocation authorization exercise separate authority. A reviewer
can certify a successor without rewriting earlier invocation records. Applying that successor to
an older call would enlarge its recorded interpreter set. A deployment with external action
approval would also enlarge the set covered by that decision. Updating the call therefore creates
a new formation record and reruns external approval when that policy is present.

This closed-target policy applies when interpreter identity, configuration, credential access, or
policy provenance forms part of approval. A deployment may instead authorize any future
implementation that satisfies a complete effect contract. That broader delegation defines a larger
formation-time authority set and does not use FCD's closed-target rule.

The enforcement invariant carries this joint decision to the effect. A fixed authenticated pool
mapping preserves the selected target identity. Mutable routing uses an effect-owning fence. The
registered source contract describes the source under covered security-relevant inputs. A manual
edge or call-specific certificate establishes successor compatibility with that contract. Action
authorization, prompt-injection resistance, and duplicate-effect prevention compose as separate
policies. The remaining design implements the rule through exact-first admission and retirement.

\subsection{Design overview and requirements}

Stable-name catalog binding implements this objective through a trusted host and gateway. It leaves
the model-visible name and arguments unchanged and resolves the retained formation binding before
backend execution (Figure~\ref{fig:drift-overview}).

\paragraph{Formation authority rule}
For invocation $i$, let $e_s$ denote the source. Let $S_i$ be empty or contain the one successor
captured at formation. Formation fixes $A_i^0=\{e_s\}\cup S_i$. Let $A_i(t)$ denote the authority
remaining after the lifecycle decisions that the deployment applies to invocation $i$ by time $t$.
Admission is the cutoff without a final guard. A deployment with a final guard extends the cutoff to
its pre-effect generation check. The remaining authority may only contract, and every protected
effect must use an authorized interpreter:
\[
\begin{aligned}
t_2\geq t_1&\Rightarrow A_i(t_2)\subseteq A_i(t_1)\subseteq A_i^0,\\
\mathsf{effect}(i,e,t)&\Rightarrow e\in A_i(t).
\end{aligned}
\]
A non-source effect also requires the source to reach \texttt{STALE}:
\[
\mathsf{effect}(i,e,t)\land e\ne e_s
\Rightarrow e\in S_i\land \mathsf{state}(e_s,t)=\texttt{STALE}.
\]
The design carries this invariant through discovery, admission, retirement, and execution.

Pinning, draining, and fencing implement identity and lifecycle eligibility. Compatibility evidence
establishes target suitability, while $A_i^0$ records invocation authority. Later evidence governs
new calls. An older call uses its captured successor or creates a new formation record and reruns
any configured action approval.

The design follows four requirements. \emph{Formation preservation} retains the catalog snapshot
used for generation and approval. \emph{Registered-target authentication} binds that snapshot to a
control-plane artifact identity. \emph{Atomic admission} orders dispatch against drain and
retirement and keeps the checked target stable. \emph{Explicit substitution} admits a replacement
through evidence that binds the source, target, tool, concrete call, and semantic contract.
When configured, action policy runs before this enforcement boundary and covers the complete
formation record.

\subsection{Catalog snapshot and authenticated context}

The prototype carries a negotiated research extension in namespaced MCP
\texttt{\_meta}~\cite{mcp-basic,mcp-versioning}. The host reserves this namespace. Both endpoints
reject conflicting negotiation, stripped credentials, and model-supplied entries.

The gateway obtains and validates the descriptors from an advertised epoch and
issues a signed credential for each tool. The trusted host removes the private
extension fields before exposing the descriptors to the model, retaining those
credentials and the sanitized descriptors in an immutable
\texttt{BoundToolCatalog}. It validates all catalog pages before publishing the
snapshot. Relisting constructs a distinct snapshot rather than replacing the
provenance of a pending invocation.

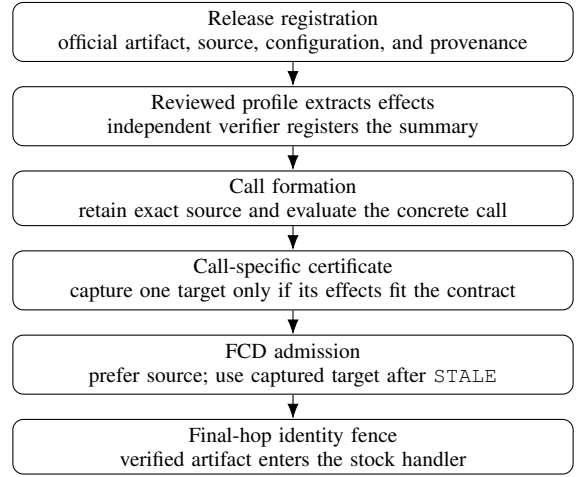
\begin{figure}[t]
\centering
\begin{tikzpicture}[
  node distance=3.2mm,
  every node/.style={font=\footnotesize,align=center},
  stage/.style={draw,rounded corners,text width=0.84\columnwidth,inner sep=3pt},
  flow/.style={-{Latex[length=2mm]}}
]
\node[stage] (release) {Release registration\\official artifact, source, configuration, and provenance};
\node[stage,below=of release] (summary) {Reviewed profile extracts effects\\independent verifier registers the summary};
\node[stage,below=of summary] (formation) {Call formation\\retain exact source and evaluate the concrete call};
\node[stage,below=of formation] (certificate) {Call-specific certificate\\capture one target only if its effects fit the contract};
\node[stage,below=of certificate] (admission) {FCD admission\\prefer source; use captured target after \texttt{STALE}};
\node[stage,below=of admission] (effect) {Final-hop identity fence\\verified artifact enters the stock handler};
\draw[flow] (release) -- (summary);
\draw[flow] (summary) -- (formation);
\draw[flow] (formation) -- (certificate);
\draw[flow] (certificate) -- (admission);
\draw[flow] (admission) -- (effect);
\end{tikzpicture}
\caption{Profile-guided evidence feeds FCD's formation and execution path. Source analysis occurs at
release registration; runtime enforcement uses the captured certificate.}
\label{fig:binding-flow}
\end{figure}

Generation and any approval retain a reference to this exact snapshot
(Figure~\ref{fig:binding-flow}). When dispatching a model-produced name and arguments, the
host obtains the credential from that snapshot, never from a mutable lookup of
the latest credential for the tool. It adds the credential to
\texttt{tools/call.params.\_meta} without changing the name or arguments. The
gateway validates and consumes the private fields, preserving unrelated
metadata, and forwards the original call to the selected stock handler.

Output-validation state is stored with each snapshot, so an earlier invocation is checked against
its own output schema after a relist.

\paragraph{Catalog and snapshot authentication}

Each credential binds service scope, route, tool name, descriptor digest, catalog epoch, artifact
identity, expiry, and signing key. The gateway obtains principal, audience, and session context
outside the model's arguments. It rejects invalid, expired, cross-context, or mismatched credentials
before dispatch, and the registry checks lifecycle state on arrival.

The trusted host associates that credential with each invocation. When enabled, an approval
callback also binds the tool-use identifier, arguments, snapshot, and captured assertion. An
optional ledger enforces one-use decisions. The base mechanism enforces formation identity without
this callback.

\subsection{Atomic admission and retirement}
\label{sec:admission}

An \texttt{ADVERTISED} epoch accepts discovery and execution.
\texttt{HIDDEN\_BUT\_VALID} removes it from discovery while retaining execution
for previously formed calls. Retirement begins at \texttt{DRAINING}: new leases
are prohibited, while admitted calls remain pinned. The epoch reaches terminal \texttt{STALE} only
after admitted and uncertain work clears. An exact-only record then rejects, while a record with a
captured successor may select that target. Redeployment creates a new epoch.

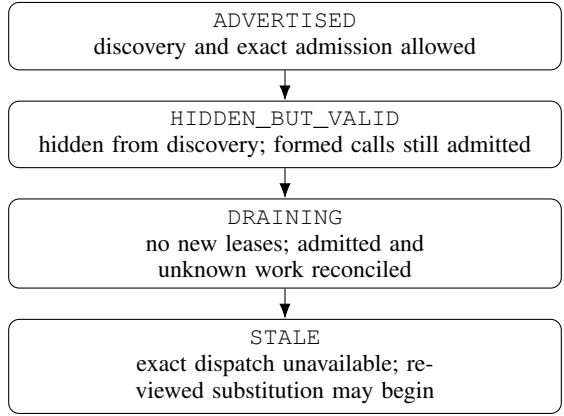
\begin{figure}[t]
\centering
\begin{tikzpicture}[
  node distance=4mm,
  every node/.style={font=\small,align=center},
  state/.style={draw,rounded corners,text width=0.82\columnwidth,inner sep=4pt},
  flow/.style={-{Latex[length=2mm]}}
]
\node[state] (advertised) {\texttt{ADVERTISED}\\discovery and exact admission allowed};
\node[state,below=of advertised] (hidden) {\texttt{HIDDEN\_BUT\_VALID}\\hidden from discovery; formed calls still admitted};
\node[state,below=of hidden] (draining) {\texttt{DRAINING}\\no new leases; admitted and unknown work reconciled};
\node[state,below=of draining] (stale) {\texttt{STALE}\\exact dispatch unavailable; reviewed substitution may begin};
\draw[flow] (advertised) -- (hidden);
\draw[flow] (hidden) -- (draining);
\draw[flow] (draining) -- (stale);
\end{tikzpicture}
\caption{Epoch lifecycle. Admission and the transition into \texttt{DRAINING} are serialized;
an expired lease alone cannot make an epoch \texttt{STALE}.}
\label{fig:epoch-lifecycle}
\end{figure}

Multiple gateway workers share a single-node SQLite registry in WAL mode.
Credential validation, lifecycle checking, and lease acquisition occur within
one \texttt{BEGIN IMMEDIATE} transaction. Thus admission and the start of drain
have an ordered outcome: admission commits and pins the target, or drain prevents dispatch. The
same lock orders credential validation, expiry, revocation, target checks, and lease creation.
Post-admission revocation is enforced by the generation checked at the final guard.

The replicated realization replaces the SQLite transaction with an etcd transaction. Admission
compares the epoch lifecycle revision, selected assertion identity, revocation-policy revision,
and target identity. The same transaction creates a unique durable marker for the selected lease.
Retirement changes the epoch to \texttt{DRAINING} with a revision compare-and-swap and reaches
\texttt{STALE} after its lease-marker prefix becomes empty. Concurrent admission therefore commits
before the drain linearization point or observes the transition and rejects. Registry
unavailability stops dispatch.

The optimized path pools etcd connections, caches verified immutable capabilities, and removes each
lease with an exact-value transaction. Retirement and reconciliation retain the full
read--validate--compare-and-swap path.

\subsection{Uncertain outcomes and recovery}

A transport return can leave an effect uncertain. FCD retains \texttt{DRAINING} and its durable
marker across coordinator failure, blocking replay and substitution. A signed resolution or newer
committed backend generation releases the lease, while the original tool use remains consumed.

The final guard reserves a signed single-use generation token before the handler and rejects replay
or stale generations. An authenticated sidecar or backend-native guard can mediate an external
effect. Our Kubernetes realization uses this fence with a three-member registry.

A stale result terminates the invocation. Recovery creates a new formation record and reruns any
configured action decision.

\subsection{Formation-bound compatibility authority}

Exact dispatch remains the default. A pending invocation can capture one successor at formation.
The capture requires evidence that names the source, target, tool, artifacts, configuration,
security contract, validity interval, sequence, and revocation handle. FCD accepts two evidence
paths. A reviewer may sign a scoped compatibility edge. A supported tool may instead use the
profile-guided call-specific path described below. Both paths fix the same target in $A_i^0$ before
execution.

Substitution begins only after the source reaches \texttt{STALE}. A \texttt{DRAINING} source may
still have an uncertain effect, so it cannot trigger fallback. The target must remain
\texttt{ADVERTISED} or \texttt{HIDDEN\_BUT\_VALID}. Admission rechecks the captured evidence,
policy sequence, revocation state, target manifest, and artifact identity before acquiring its
lease. Temporary source unavailability causes rejection or exact retry. It does not authorize the
successor.

The host can capture an exact successor identity and certificate before that target becomes
admission-eligible. A later registry entry activates only the identity already present in $A_i^0$.
The source still receives priority. If it completes retirement before execution, the gateway may
select the captured target. A certificate first issued after formation governs later formations.
A call that captured v2 cannot acquire v3 through a later v2--v3 decision.

\subsection{Profile-guided call-specific evidence}
\label{sec:effect-profiles}

A whole release can be incompatible while a particular call remains safe. A repository handler may
change the default for an omitted \texttt{private} field, while explicit
\texttt{private:true} retains the same effect. FCD therefore evaluates compatibility for the
concrete call before it captures a successor.

A reviewed \emph{effect profile} defines five items for one handler path: accepted source structure,
supported arguments, modeled effects, external assumptions, and rejection conditions. The profile
also states its completeness boundary. Each language frontend emits the same intermediate effect
record. That record names a domain, operation, principal, destination, resource, exposure level,
mutation bit, and write scopes. Our four frontends cover a Python process sink, a JavaScript SQL
policy, a Go HTTP request, and a C\# cloud configuration path.

The trusted host or an external action-policy layer creates the formation contract $C_i$. It binds
the descriptor and approval to the concrete principal, operation, destinations, resource prefixes,
maximum exposure, mutation permission, and write scopes. This contract supplies user policy. The
profile computes the effects that one interpreter may produce for the formed arguments. The registrar
establishes that the source call fits $C_i$. Successor analysis checks the target against the same
contract.

Registration binds the official artifact to analyzed source and release provenance. The registrar
authenticates this relation as a trusted provenance assertion. Reproducible builds or in-toto-style
attestations can provide stronger evidence for this assertion~\cite{in-toto}. Reproducible
compilation remains a separate supply-chain property. The common pipeline then applies the
following procedure:
\begin{enumerate}
  \item The profile matcher locates the reviewed handler path and emits source facts with line and
  hash anchors.
  \item The call-site guard extracts callee identities from each declared handler region. It rejects
  identities outside the profile's reviewed inventory and binds the result to the source.
  \item The summarizer maps the source facts to an ordered first-match rule set. Each rule produces an
  effect record, handler rejection, or \texttt{UNKNOWN}.
  \item A second process regenerates the facts and rules from the registered source. Matching
  profile, artifact, configuration, summary, closure, and evidence digests admit the summary to the
  registry.
  \item Formation evaluates the rules with the concrete arguments and authenticated environment.
  It issues a certificate only when every emitted effect satisfies $C_i$.
\end{enumerate}
This interface separates reusable certificate and containment logic from each language-specific
source matcher. A recognized callee outside the reviewed inventory stops registration before the summary can
authorize execution.

\begin{table}[t]
\centering\scriptsize
\setlength{\tabcolsep}{2.5pt}
\caption{Bounded call-site inventories. Each row compares the recognized callee set in one source
region with a reviewed allowlist.}
\label{tab:profile-call-inventories}
\begin{tabularx}{\columnwidth}{@{}>{\raggedright\arraybackslash}p{0.20\columnwidth}>{\raggedright\arraybackslash}p{0.36\columnwidth}X@{}}
\toprule
Profile & Reviewed region & Inventory item \\
\midrule
Reference Git / Python & AST body of \texttt{git\_diff} & Resolved callee path \\
DBHub / JavaScript & Four named classifier and handler functions & Dotted callee token \\
GitHub / Go & Named \texttt{CreateRepository} block & Dotted callee token \\
Azure / C\# & Unique \texttt{CreateResourceAsync} method & Dotted callee token \\
\bottomrule
\end{tabularx}
\end{table}

The inventory stores callee identities rather than source positions or call counts. The Python
frontend resolves AST call paths. The other frontends extract dotted callee tokens from their
bounded source spans. The DBHub span covers \texttt{checkReadOnly}, \texttt{isReadOnlySQL},
\texttt{areAllStatementsReadOnly}, and \texttt{createExecuteSqlToolHandler}. A missing or ambiguous
region rejects registration. A dynamic Python callee also rejects registration. The source matcher
and summarizer separately analyze supported semantic facts and map them to effect rules. Passing the
guard is a registration prerequisite. The guard itself grants no execution authority. The reviewed
envelope defines the region, dependencies, environment, and modeled effects.

\paragraph{Complete GitHub derivation}
The GitHub profile makes this pipeline concrete. The matcher accepts one
\texttt{CreateRepository} handler and locates the parser for \texttt{private}. It then verifies the
flow through \texttt{github.Ptr(private)} to \texttt{client.Repositories.Create}. The v1.4 source
fact records that omission defaults to \texttt{true}; the v1.3 fact records \texttt{false}. The
summary converts this fact into the rule
\texttt{private := arg\_default(private, default)} and maps the Boolean value to private or public
exposure. Consider the concrete call \texttt{\{name: project-x\}}. Its contract permits repository
creation for \texttt{project-x} at the authenticated GitHub destination and sets the maximum
exposure to private. The v1.4 summary therefore admits the omitted-field call. The v1.3 summary
produces public exposure and rejects that same call. Explicit \texttt{private:true} remains inside
the contract under both summaries. The profile covers this parser-to-request path. Its registered
inputs include the client library, account, organization, API base URL, and middleware. The effect
ends at the authenticated repository-creation request.

Formation applies the target summary to the concrete arguments and authenticated execution context.
Let $C_i$ denote the call's formation-time security contract. Let $S(e,i)$ denote the generated
over-approximation of in-scope effects for interpreter $e$ and invocation $i$. The verifier captures
target $e_t$ only when
\[
  \mathsf{ActualInScopeEffects}(e_t,i) \subseteq S(e_t,i) \subseteq C_i.
\]
The first inclusion is the reviewed profile premise. It states that the profile over-approximates
the target's effects within its declared source, argument, dependency, and environment envelope.
The common verifier establishes the second inclusion from the rule evaluation and $C_i$. The
certificate validates summary containment under this premise. Profile correctness remains the
reviewed first-inclusion premise. The certificate binds the arguments, principal, destination,
source and target manifests, profile,
summary, contract, and configuration digests. Its target then enters $A_i^0$. This property is
contract containment. It permits target behavior that differs from the source while remaining
inside the authority approved for the call.

Execution authority comes from the verified certificate captured by FCD. A missing source fact,
unsupported argument, ambiguous structure, provenance mismatch, or unknown effect stops automatic
capture. The host can continue with exact-only FCD, obtain a manually reviewed edge, or form a new call after the
update. This rule lets one release preserve safe calls without treating every call as compatible.

The analysis runs before a release becomes admission-eligible. Registration verifies the frozen
evidence and exact target identity. The dispatch path verifies the captured certificate and uses
the existing admission, lease, revocation, and final-fence checks. Generated and manual evidence
therefore share one runtime enforcement boundary.

\section{Security Analysis}
\label{sec:security-analysis}

We analyze accepted dispatches under Section~\ref{sec:invariant}'s trust boundary. The host and
gateway use unforgeable credentials over an integrity-protected channel. The registry mediates every
pool and maps each registered identity to its artifact. The lifecycle authority resolves admitted
and uncertain work before declaring a source \texttt{STALE}.

\paragraph{Proposition: formation-authorized effect}
Every protected effect for invocation $i$ uses an interpreter in $A_i^0$ when P1--P4 hold. A
non-source effect occurs only after the source reaches \texttt{STALE}. P1 retains formation
provenance. P2 prevents later authority expansion. P3 orders admission against lifecycle changes.
P4 carries the admitted identity to the effect-owning path.

\subsection{P1: Formation provenance}

The host retains the immutable snapshot that shaped each invocation. Its credential authenticates
the descriptor, epoch, tool, artifact, and session. The gateway rejects a different binding before
dispatch. Relisting and reconnection therefore cannot replace formation provenance. Faithful
registration connects the artifact to the model-visible contract.

\subsection{P2: Monotonic target authority}

Formation fixes the source and at most one captured target. That target carries either a reviewed
edge or a verified call-specific certificate. It may register or become runnable later, but its
identity cannot change. Expiry and revocation before admission remove it. They cannot add a
replacement. A later certificate governs new invocations. Reusing the same arguments with a new
target requires a new formation record and any configured action approval.

FCD defines two revocation cutoffs. A revocation ordered before admission prevents lease creation.
An admitted lease wins the registry order. Deployments that apply post-admission revocation carry
the checked generation to the final guard. The guard rechecks it immediately before effect. The
authoritative policy clock gives every gateway replica the same expiry and revocation order.

\subsection{P3: Atomic lifecycle admission}

The registry checks binding and lifecycle state in the transaction that acquires a lease. Admission
pins the selected pool when it commits first. Drain or revocation rejects the call when its
transaction commits first. A \texttt{DRAINING} source cannot trigger substitution.

SQLite supplies this order on one host. The replicated implementation uses etcd transactions. Each
admission compares the lifecycle revision, certificate identity, revocation-policy revision, and
target identity before creating a unique lease marker. Retirement changes the epoch to
\texttt{DRAINING}, which fences new markers. It reaches \texttt{STALE} only after the marker prefix
becomes empty. We exercised this order on a kind cluster and across three EC2 guest VMs.

\subsection{P4: Execution identity}

An accepted lease names the pool and artifact checked at admission. A fixed authenticated mapping
preserves this relation when every member proves the registered identity. A mutable route instead
carries a signed generation to a non-bypassable final guard. The guard rejects stale, mismatched,
or replayed tokens before the effect.

Our Kubernetes fixture owns the recorded effect and checks artifact and configuration identity on
that path. An external service can place the check in a backend-native guard or authenticated
sidecar that mediates the only effect route. Workload identity authenticates the component that
hosts the guard.

\subsection{P5: Profile-bounded effect containment}

A generated certificate adds a semantic premise to P1--P4. The trusted host constructs $C_i$ from
the descriptor, concrete arguments, authenticated principal and destination, and any external
action approval. The registrar establishes that the source call satisfies this contract. For each
supported profile, the reviewed envelope over-approximates the actual in-scope effects of accepted
source structures, arguments, dependencies, and environments. The independent verifier regenerates
the target summary from provenance-bound source. It also checks each callee identity recognized by
the language frontend against the reviewed inventory for that handler region. Formation captures the
target only when that summary is contained by $C_i$. Therefore
\[
\mathsf{ActualInScopeEffects}(e_t,i)\subseteq S(e_t,i)\subseteq C_i.
\]
The certificate checks the second inclusion. The reviewed acceptance envelope supplies the first
inclusion as a trusted premise. Certificate verification establishes profile-bounded containment
under that premise.
P1--P4 ensure that execution reaches the same target and environment named by this certificate.
P5 then gives contract containment for the protected effect. Its semantic claim is the formation
authority expressed by $C_i$ for this call.

Profile authorship defines the supported semantics. The profile records its argument domain,
external assumptions, reviewed regions, call inventory, and completeness boundary. Source-shape
mismatch, an unreviewed call site, unsupported arguments, unknown effects, and identity mismatch stop
the generated path. A manually reviewed edge uses the same lifecycle but retains the reviewer's
compatibility judgment as its semantic premise.

Together, P1--P4 establish formation-authorized registered-target safety. P5 establishes
call-specific semantic containment within a reviewed profile. Availability follows the remaining
authorized set. The call ends when that set contains no eligible interpreter.

\section{Evaluation}
\label{sec:evaluation}

\subsection{Prototype and setup}

The prototype uses public interfaces from the official Python MCP SDK. Its local implementation
stores immutable host snapshots and lifecycle state in SQLite WAL. Its replicated implementation
uses two gateway Pods, a three-member etcd registry, and one marker per admitted lease. A durable
continuation store restores formation records after reconnect. The final guard verifies artifact,
configuration, and generation identity before the effect.

The profile pipeline binds official source and release provenance to executable artifacts. It uses
four reviewed analyzers: Python AST for a process sink, bundled JavaScript structure for SQL policy,
Go source structure for an HTTP request, and C\# source structure for cloud configuration. A second
process regenerates every summary before registration. The call verifier substitutes concrete
arguments and authenticated context, checks containment, and issues the certificate consumed by
FCD.

Local experiments used macOS arm64, Python 3.11, SQLite WAL, isolated stock handlers, fake
credentials, and local effect recorders. Agent paths covered LangGraph, LangChain MCP, PydanticAI,
and the Python MCP SDK. Replicated tests used kind and three-node k3s deployments with etcd 3.5.17;
the EC2 runs used a separate load generator in one availability zone. Model experiments used
\texttt{qwen3:8b}, \texttt{gpt-5.6-sol}, Nova Lite, and Gemini 2.5 Flash Lite. The evidence package
freezes exact versions, source hashes, schedules, requests, effects, and recomputation code.

The evaluation asks four questions. RQ1 measures profile-guided evidence generation and its
connection to stock effects. RQ2 tests formation authority and lifecycle enforcement. RQ3 tests the
same boundary across frameworks and deployment paths. RQ4 measures model-facing agreement and
runtime cost.

\subsection{RQ1: Profile-guided compatibility evidence}

We evaluated four profiles over 32 official releases. The first cohort covers Reference Git and
DBHub over 16 releases. Thirteen releases generated summaries that agreed with stock-effect
endpoints. Three DBHub releases moved the classifier across module boundaries, left the reviewed
source shape, and escalated. The frozen expansion adds GitHub
and Azure profiles. We built the profiles from two development releases per product, froze them,
and selected 16 other releases from commit metadata. All 16 generated without profile changes and
matched a separately implemented source oracle. Table~\ref{tab:profile-results} keeps the two
reference types separate.

\begin{table}[t]
\centering\scriptsize
\setlength{\tabcolsep}{2pt}
\caption{Security-relevant regimes across official releases. Repeated releases within a regime
measure profile reuse; each arrow marks a change in the concrete-call decision.}
\label{tab:profile-results}
\begin{tabularx}{\columnwidth}{@{}>{\raggedright\arraybackslash}p{0.20\columnwidth}>{\raggedright\arraybackslash}p{0.23\columnwidth}X>{\raggedright\arraybackslash}p{0.16\columnwidth}@{}}
\toprule
Profile & Release groups & Modeled transition & Check \\
\midrule
Reference Git / Python & 4 unsafe + 4 reject & Dash target: \texttt{UNSAFE}$\rightarrow$\texttt{REJECT} & Stock 8/8 \\
DBHub / JavaScript & 4 unsafe + 1 reject + 3 escalation & PRAGMA assignment: \texttt{UNSAFE}$\rightarrow$\texttt{REJECT}; later source-shape exit & Stock 5/5 \\
GitHub / Go & 4 default-false + 4 default-true & Omitted \texttt{private}: \texttt{UNSAFE}$\rightarrow$\texttt{SAFE} & Oracle 8/8 \\
Azure / C\# & 4 option-absent + 4 option-wired & Omitted network option: \texttt{UNSAFE}$\rightarrow$\texttt{SAFE} & Oracle 8/8 \\
\midrule
Total & 29 automatic + 3 escalation & 4 modeled transitions; 3 structure exits & 13 stock + 16 oracle \\
\bottomrule
\end{tabularx}
\end{table}

Each profile produced two security-relevant regimes. Repeated releases within a regime retained the
same source facts and call decisions, which measures release-level reuse. The boundary between
regimes changed the generated facts and the concrete-call decision. DBHub then left the reviewed
source structure, and the generator escalated those three releases. For GitHub, explicit
\texttt{private:true} remained \texttt{SAFE} on both sides of the default change. Only the omitted
call changed its decision. The older Azure regime did not consume the network option and returned
\texttt{UNKNOWN} for explicit values. The wired regime classified explicit disabled as \texttt{SAFE}
and explicit enabled as \texttt{UNSAFE}.

Representative boundaries are Reference Git \texttt{2025.9.25}$\rightarrow$\texttt{2026.1.14},
DBHub \texttt{0.22.5}$\rightarrow$\texttt{0.23.0}, GitHub
\texttt{v1.2.0}$\rightarrow$\texttt{v1.5.0}, and Azure
\texttt{beta.30}$\rightarrow$\texttt{beta.33}. DBHub \texttt{0.24.0} begins the source-shape exit.

In this transition-centered reuse study, the existing profiles processed 29 of 32 releases without
release-specific edits. The remaining three releases left their reviewed source structures and
escalated. This ratio describes reuse within the selected transitions.

The expansion freeze preceded holdout-source inspection. It fixed profile hashes, calls,
classifications, and 16 releases selected from tag and commit metadata. A separate oracle checked
runtime-value assignment and effect sinks without using generated facts or templates. Both new
profiles processed eight unseen releases without modification. A later fixture repair replaced a
synthetic Azure \texttt{name} with the stock \texttt{resource} argument; the reported call-level
result uses that repaired run, while the original freeze supplies release-level holdout counts.

The generated evidence remained call-specific. Across the frozen expansion, the verifier preserved
all 20 supported safe calls. It returned \texttt{UNKNOWN} for an explicit Azure option that an old
handler did not consume.

Summary regeneration checks that a source and profile reproduce the registered rules. We added the
call-site guard after completing the two frozen release holdouts. We then preregistered a separate
regression study over their retained sources. The study fixed the 32-release denominator, four
reviewed inventories, four representative sources, and 16 mutation conditions. The guard preserved
all 29 previously automatic release decisions and introduced no new escalation. The three
unsupported DBHub releases retained their earlier escalation.

We added filesystem, network, and process call sites to one representative official source per
profile. The guard rejected all 12 injected additions within the reviewed regions. It accepted all
four comment-only controls. The release rows measure compatibility with prior automatic decisions.
The mutation rows measure detection for the three added call classes. The source matcher and
summarizer retain responsibility for semantic changes among reviewed calls.

We connected one representative release from each profile to the complete protected path. Reference
Git \texttt{2025.9.25} and DBHub \texttt{0.22.5} executed safe calls through their unmodified stock
handlers. Their dangerous calls stopped at admission. GitHub v1.4.0 created private intent for the
safe call and blocked explicit public intent. Azure beta.32 produced disabled-network intent for
the safe call and blocked enabled-network intent. Unprotected controls produced the corresponding
canary file, database mutation, public repository intent, and enabled-network intent.

Each trace began with official source and artifact provenance. It then regenerated the summary,
formed a call-specific certificate, acquired durable admission, rechecked workload identity, and
entered the stock handler. Replacing the registered artifact caused
\texttt{final\_hop\_artifact\_mismatch}. The mismatched path invoked zero handler callbacks. Local
mock endpoints recorded the GitHub and Azure requests without contacting production services.

\begin{table}[t]
\centering\scriptsize
\setlength{\tabcolsep}{2.5pt}
\caption{Semantic mutations mapped to reviewed proof obligations. All seven produced the expected
result.}
\label{tab:semantic-mutations}
\begin{tabularx}{\columnwidth}{@{}>{\raggedright\arraybackslash}p{0.27\columnwidth}>{\raggedright\arraybackslash}p{0.43\columnwidth}X@{}}
\toprule
Mutation & Proof obligation & Observed result \\
\midrule
Git comment spoof & Comments cannot create a guard fact & Dangerous state retained \\
Git guard after sink & Dash rejection must precede the sink & Dangerous state retained \\
Git missing sink & One direct diff sink is required & Registration rejected \\
DBHub comment spoof & Comments cannot create an assignment guard & Dangerous state retained \\
DBHub missing guard & Read-only guard is required & Registration rejected \\
DBHub fall-through guard & Guard failure must return before the sink & Registration rejected \\
DBHub duplicate classifier & Classifier structure must be unique & Registration rejected \\
\bottomrule
\end{tabularx}
\end{table}

Table~\ref{tab:profile-envelopes} exposes the four reviewed acceptance envelopes.
Table~\ref{tab:semantic-mutations} connects the seven existing mutations to conditions within the
Reference Git and DBHub envelopes. The profile author supplies the first inclusion in
$\mathsf{ActualInScopeEffects}\subseteq S\subseteq C_i$. These tables make its scope and tested
obligations explicit. All tests in the current reproducibility capsule passed.

RQ1 shows that reviewed profiles reuse source-derived effect evidence while preserving explicit
escalation. The call-site guard detects the tested additions to the reviewed inventory. Stock traces
connect generated facts to admission and effect, while the envelope and mutation tables expose the
reviewed premise. RQ2 composes this evidence with formation authority and source retirement through
the same admission and final-fence boundary.

\subsection{RQ2: Formation authority and lifecycle}

We first test whether call-specific evidence preserves useful pending work. Both calls formed and
received approval under stock GitHub MCP v1.4.0. Call A explicitly set
\texttt{private:true}, which remains private under v1.3.0. Call B omitted \texttt{private}, which
becomes public under v1.3.0. After v1.4 retirement, the preregistered comparison applied four
policies to three repetitions of each call (Table~\ref{tab:policy-granularity}).

\begin{table}[t]
\centering\footnotesize
\setlength{\tabcolsep}{2.4pt}
\caption{Call granularity after source retirement. A remains private on the target; B becomes
public. ``New'' counts calls that require new formation and approval.}
\label{tab:policy-granularity}
\begin{tabularx}{\columnwidth}{@{}>{\raggedright\arraybackslash}p{0.29\columnwidth}rrrr@{}}
\toprule
Policy & A & B & New & Violation \\
\midrule
Exact pinning & 0/3 & 0/3 & 6 & 0 \\
Release-wide deny & 0/3 & 0/3 & 6 & 0 \\
Release-wide allow & 3/3 & 3/3 & 0 & 3 public \\
Call-specific FCD & 3/3 & 0/3 & 3 & 0 \\
\bottomrule
\end{tabularx}
\end{table}

Exact pinning preserved safety by requiring new formation and approval for all six calls. The two
release-wide policies either stopped both call classes or admitted three public effects. FCD
completed all three A calls under their existing formation and approval. It stopped all three B
calls before the stock handler callback and required new approval only for B. An independent
auditor reconstructed all four rows from raw records and passed 12/12 gates. All 27 regression tests
passed, and the repeated audit was byte-identical.

We next isolate when a successor acquires authority. One frozen stock-MCP experiment compared
exact pinning, execution-time adoption, a captured manual edge, and the same edge installed after
formation. A second frozen experiment replaced the manual edge with generated call-specific
evidence and added a fresh-formation control. Both used unmodified GitHub MCP v1.4.0 and v1.5.0,
identical runtime descriptors, distinct artifact identities, a retired and paused source, and
separate effect recorders. Table~\ref{tab:compatibility} keeps their denominators separate.

\begin{table}[t]
\centering\footnotesize
\setlength{\tabcolsep}{2.5pt}
\caption{Successor authority after source retirement. Manual-edge and generated-certificate rows
come from separate frozen stock experiments.}
\label{tab:compatibility}
\begin{tabularx}{\columnwidth}{@{}>{\raggedright\arraybackslash}p{0.27\columnwidth}>{\raggedright\arraybackslash}p{0.31\columnwidth}X@{}}
\toprule
Condition & Pending authority & Outcome after retirement \\
\midrule
Exact pinning & Source only & Reject, 0/5 effects \\
Execution-time adoption & Late manual edge rewrites record & v1.5 effect, 5/5 \\
Captured manual edge & v1.5 captured at formation & v1.5 effect, 5/5 \\
Late manual edge & Source-only record retained & Reject, 0/5 effects \\
Captured generated cert. & v1.5 identity and cert. captured & v1.5 effect, 3/3 \\
Late generated cert. & Old record remains source-only & Old callback 0/3; fresh formation 3/3 \\
\bottomrule
\end{tabularx}
\end{table}

The execution-time control rewrote pending authority and reached v1.5. FCD retained the original
record and rejected the same late edge. The matched pair isolates formation-time capture from
signature validity and target identity.

The generated-certificate experiment then joined semantic evidence to this lifecycle. A certificate
captured at formation executed v1.5 after v1.4 retirement in 3/3 trials. The same safe certificate
installed later produced zero old-call callbacks, while a fresh formation executed v1.5 in 3/3.
Explicit public intent and a mismatched final artifact each stopped before handler effect in 3/3.
An independent auditor recomputed all 12 outcomes. Additional signed-edge and authored-transition
tests matched their target and effect oracles. A breaking transition retained an unchanged
descriptor, so descriptor equality did not recover formation authority.

A captured candidate also needs lifecycle enforcement. A matched ablation compared FCD with a
formation-time fallback that selected its successor whenever the source route disappeared. With an
active source lease, the fallback executed B in 20/20 schedules while FCD rejected before effect in
20/20. After admission, a mutable A$\rightarrow$B route reached B under the fallback in 20/20
schedules. FCD's A-bound token stopped at the final guard in all 20. Safe-retirement and stable-route
controls reached B and A respectively in 20/20 schedules.

Lifecycle tests covered acquire--drain races, reconnect, worker death, registry loss, shared
coordination, backend fencing, policy-clock skew, and unknown outcomes. Every admitted call produced
at most one authorized effect. Unknown outcomes remained \texttt{DRAINING} until signed resolution,
and the original tool-use record remained consumed. Across local and EC2 deployments, seven
quorum-loss and 25 post-formation drain schedules stopped before effect.

These results answer RQ2: under the closed-target policy, effect compatibility does not itself
authorize a target for an existing call. Formation capture determines successor authority.
Completed retirement, transactional admission, and final-hop identity checks determine whether
that authority can produce an effect.

\subsection{RQ3: Integration and deployment transfer}

We inserted the trusted bridge into hosted LangGraph, native LangChain MCP, PydanticAI, process
continuation, and replicated Kubernetes paths. Current epochs reached the exact handler. Draining
or stale epochs stopped before handler effect or model continuation. Stateless adapters opened new
sessions between discovery and execution, so the bridge restored the authenticated formation record
rather than relying on connection affinity. A process handoff restored the same record in a new
process and rejected replay.

The local Kubernetes experiment moved formation and execution between two gateway replicas. Its 64
schedules covered cross-replica execution, same-image configuration drift, quorum loss, and drain.
Every schedule produced one registered effect or a pre-effect rejection. The three-VM EC2 deployment
repeated 18 schedules. Five cross-replica calls reached the registered target. Five configuration
mismatches, three quorum-loss schedules, and five drains rejected before effect. A paired Azure
release path also preserved disabled-network intent under exact execution and blocked the pending
call during drain in 3/3 trials.

The same authenticated formation record and shared admission order governed every tested framework,
process-handoff, and cross-replica path.

\subsection{RQ4: Model-facing agreement and runtime cost}

Hidden metadata preserved the model-visible interface. Three paired 100-prompt experiments over 19
GitHub tools produced FCD/plain accepted-call counts of 64/63 for GPT, 65/64 for Nova Lite, and
74/74 for Gemini Flash Lite. Each prompt used a fresh conversation and one frozen task oracle.
Exact McNemar tests gave $p=1.00$ for GPT and Nova; Gemini had no discordant pairs. Every accepted
protected call used the registered identity.

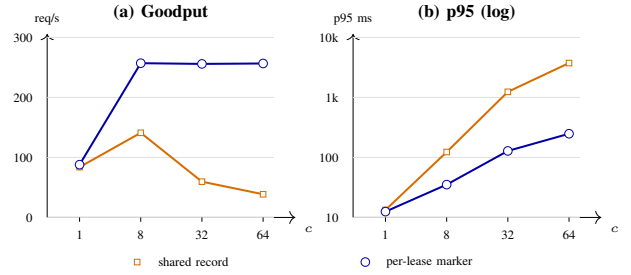
\begin{figure}[t]
\centering
\resizebox{0.96\columnwidth}{!}{%
\begin{tikzpicture}[font=\tiny]
  \begin{scope}
    \draw[->] (0,0) -- (3.45,0) node[below right] {$c$};
    \draw[->] (0,0) -- (0,2.55) node[above] {req/s};
    \foreach \y/\lab in {0/0,0.833/100,1.667/200,2.5/300} {
      \draw[black!12] (0,\y) -- (3.15,\y);
      \node[anchor=east] at (-0.08,\y) {\lab};
    }
    \foreach \x/\lab in {0.45/1,1.3/8,2.15/32,3.0/64}
      \draw (\x,0) -- +(0,-0.06) node[below] {\lab};
    \draw[orange!85!black,thick]
      (0.45,0.699) -- (1.3,1.176) -- (2.15,0.497) -- (3.0,0.322);
    \foreach \x/\y in {0.45/0.699,1.3/1.176,2.15/0.497,3.0/0.322}
      \node[rectangle,draw=orange!85!black,fill=white,inner sep=1.1pt] at (\x,\y) {};
    \draw[blue!65!black,thick]
      (0.45,0.733) -- (1.3,2.143) -- (2.15,2.133) -- (3.0,2.139);
    \foreach \x/\y in {0.45/0.733,1.3/2.143,2.15/2.133,3.0/2.139}
      \node[circle,draw=blue!65!black,fill=white,inner sep=1.2pt] at (\x,\y) {};
    \node[font=\scriptsize\bfseries] at (1.58,2.85) {(a) Goodput};
  \end{scope}

  \begin{scope}[xshift=4.25cm]
    \draw[->] (0,0) -- (3.45,0) node[below right] {$c$};
    \draw[->] (0,0) -- (0,2.55) node[above] {p95 ms};
    \foreach \y/\lab in {0/10,0.833/100,1.667/1k,2.5/10k} {
      \draw[black!12] (0,\y) -- (3.15,\y);
      \node[anchor=east] at (-0.08,\y) {\lab};
    }
    \foreach \x/\lab in {0.45/1,1.3/8,2.15/32,3.0/64}
      \draw (\x,0) -- +(0,-0.06) node[below] {\lab};
    \draw[orange!85!black,thick]
      (0.45,0.104) -- (1.3,0.909) -- (2.15,1.745) -- (3.0,2.145);
    \foreach \x/\y in {0.45/0.104,1.3/0.909,2.15/1.745,3.0/2.145}
      \node[rectangle,draw=orange!85!black,fill=white,inner sep=1.1pt] at (\x,\y) {};
    \draw[blue!65!black,thick]
      (0.45,0.081) -- (1.3,0.456) -- (2.15,0.924) -- (3.0,1.163);
    \foreach \x/\y in {0.45/0.081,1.3/0.456,2.15/0.924,3.0/1.163}
      \node[circle,draw=blue!65!black,fill=white,inner sep=1.2pt] at (\x,\y) {};
    \node[font=\scriptsize\bfseries] at (1.58,2.85) {(b) p95 (log)};
  \end{scope}

  \node[rectangle,draw=orange!85!black,fill=white,inner sep=1.1pt] at (1.25,-0.62) {};
  \node[anchor=west] at (1.42,-0.62) {shared record};
  \node[circle,draw=blue!65!black,fill=white,inner sep=1.2pt] at (4.42,-0.62) {};
  \node[anchor=west] at (4.59,-0.62) {per-lease marker};
\end{tikzpicture}%
}
\caption{Registry data model under replicated load. The per-lease run sustains higher goodput and
lower tail latency as concurrency $c$ increases. Points are medians over five repetitions.}
\label{fig:distributed-scaling}
\end{figure}

Local binding added approximately 0.030\,ms at p50 and p95. The final fence added
0.321/0.431\,ms at p50/p95. The first replicated registry updated one shared epoch record and
collapsed under contention. At concurrency 64 it sustained 38.56 requests/s with a
3,753.11\,ms p95. Unique per-lease markers sustained 256.72 requests/s with a 248.42\,ms p95
(Figure~\ref{fig:distributed-scaling}). The per-lease run avoided the shared-record write contention
observed in the first design.

The optimized EC2 path completed 1,920 measured requests across three concurrency levels and five
paired repetitions. Sequential dispatch added 10.42\,ms at p50. At concurrency 32, protected p50
was 182.31\,ms, p95 was 211.98\,ms, and goodput reached 62.7\% of the matched TLS baseline. All 1,200
protected requests, including warm-up traffic, reached the registered fixture without a drifted
effect. Profile analysis occurs at release registration, so these measurements cover the complete
runtime certificate, admission, registry, audit, and fence path.

\section{Discussion}
\label{sec:discussion}

\begin{table*}[t]
\centering
\footnotesize
\setlength{\tabcolsep}{3pt}
\renewcommand{\arraystretch}{1.05}
\caption{Closest mechanisms under one pending-work trace.}
\label{tab:related-comparison}
\begin{tabularx}{\textwidth}{@{}>{\raggedright\arraybackslash}p{0.16\textwidth}>{\raggedright\arraybackslash}p{0.19\textwidth}>{\raggedright\arraybackslash}p{0.19\textwidth}>{\raggedright\arraybackslash}p{0.20\textwidth}X@{}}
\toprule
Mechanism & Protected object & While source exists & Target first authorized after formation & After source retirement \\
\midrule
Exact pinning & Target version & Execute exact source & Reject pending call & Reject \\
Execution-time compatibility & Current compatibility policy & Select current target & May authorize target for pending work & Select under current policy \\
Formation-time fallback & Captured candidates & Prefer exact source & Reject for pending call & Select fallback; lifecycle unspecified \\
Semantic differencing~\cite{symdiff,idise} & Cross-version behavior & Compare implementations & Analyze later target & No invocation authority \\
ETDI~\cite{etdi} & Approved definition version/hash & Use approved definition & Require reapproval & No call-specific retirement rule \\
Attested admission~\cite{attested-admission} & Selected server & Authenticate server & Authenticate later server & No formation-scoped successor \\
Tool Forge~\cite{tool-forge} & Validated execution capsule & Execute with evidence & Constrain to evidence & No source-retirement rule \\
Cordon~\cite{cordon} & Staged task effects & Validate before commit & Revalidate transaction & Task-level commit authority \\
\textbf{FCD} & \textbf{Interpreter set and effect contract} & \textbf{Execute exact source} & \textbf{Require new formation} & \textbf{Execute captured target if call effects fit} \\
\bottomrule
\end{tabularx}
\end{table*}

\subsection{Authority beyond version pinning}

Table~\ref{tab:related-comparison} separates target identity from call-specific suitability and
invocation authority. Table~\ref{tab:policy-granularity} shows the operational consequence: FCD
retains existing approval for the safe call, while release-wide policies either reject both calls
or admit the public effect. Table~\ref{tab:compatibility} isolates authorization timing by comparing
a captured certificate with the same certificate installed later.

The closed-target rule applies when interpreter identity, configuration, credential access, or
policy provenance is part of approval. A deployment can instead delegate at formation to every
future implementation satisfying a complete effect contract. That policy grants broader authority
and does not require target closure.

A planned rollout illustrates the closed-target policy. During a v1--v2 overlap, a security team
approves the exact v1 and v2 artifact and configuration identities. A compatibility issuer separately
certifies that v2 satisfies the concrete call's effect contract. Formation captures v2 in the
pending record before the action enters a delayed approval or durable continuation. If v1 completes
retirement first, the captured v2 can execute the call without reconstructing the workflow or
rerunning its configured action approval. A later v3 or certificate cannot inherit that record. New
formations use the then-current policy.

The evaluated prototype records one successor. A finite ordered successor set is future work and
can preserve the same rule by fixing membership and order at formation.

\subsection{Semantic trust boundary}

Profile authors define supported structures, arguments, effects, and assumptions. The manifest
binds artifacts, configuration, flags, destinations, and dependencies, while the final guard checks
the same registered inputs. Unsupported source shapes and arguments stop automatic capture. An
independent process regenerates every summary. The call-site guard compares callee identities
recognized inside each declared region with its reviewed inventory. Appendix~\ref{app:registration}
describes this boundary.

Profile authors and the reviewed call-site classifier define the semantic TCB. P5 covers the
declared regions, arguments, effects, and external assumptions. P1--P4 preserve registered-target
identity independently of this semantic premise. Manual edges use reviewer judgment as an
alternative premise.

\subsection{Prompt and model controls}

Prompt policies can materialize known fields but cannot select the handler that receives a call.
In our control, prompt-only calls used safe explicit arguments, yet nine of 16 completed calls
reached a release different from the visible schema. Schema rereading produced no wire-level relist
in eight trials. FCD instead authenticates formation identity outside the model interface. Model
policy and dispatch integrity compose by binding any action approval to the same formation record.

\subsection{Cost, availability, and recovery}

Local enforcement adds sub-millisecond work. Replicated admission adds registry and final-fence
latency before model and external-tool execution. Section~\ref{sec:evaluation} reports both costs.

Registry outage suspends admission. Retirement, expiry, mismatch, or unresolved timeout ends or
quarantines the call rather than redirecting it. Operators can retain a source as
\texttt{HIDDEN\_BUT\_VALID}, rate-limit lifecycle changes, separate rollout and retirement keys,
and alert on epoch churn.

\section{Related Work}
\label{sec:related}

Table~\ref{tab:related-comparison} separates program evidence from authority over a pending call.
FCD derives a concrete-call effect bound, captures target authority at formation, and carries that
decision through source retirement.

\subsection{API evolution and update consistency}

Protocol Buffers and Google AIP-180 distinguish wire, source, and semantic compatibility
~\cite{protobuf-evolution,aip-180}; Web API research reduces client disruption
~\cite{schmiedmayer-webapi}. Semantic differencing and change-impact analysis compare program
behavior or affected paths~\cite{symdiff,idise}. FCD uses a narrower product: a conservative
security-effect bound for one reviewed handler path and concrete call. It then binds that evidence
to invocation authority and artifact identity.

Dynamic-update and upgrade-testing systems coordinate changed state or detect incompatible formats
~\cite{ajmani-upgrades,morpheus,zhang-upgrade,upfuzz}. Per-packet consistency retains an ingress
configuration, while Temporal binds workflows to build history or a rollout target
~\cite{reitblatt-updates,temporal-worker-versioning}. Rebound orders rollback and de-authorization
~\cite{rebound}. Routing, workload identities, leases, snapshots, draining, and fencing preserve
selected identity and lifecycle order~\cite{istio-routing,spiffe-standard,chubby,kitsune,postgres-mvcc}.
FCD uses these mechanisms as its execution substrate. Its profile decides whether a target satisfies
the concrete call's effect contract. Its formation record decides whether that invocation authorized
the target. The resulting mismatch is relational TOCTOU, enforced at admission rather than replayed
after a database race~\cite{cwe-toctou,racedb}.

\subsection{Agent and tool defenses}

Parasites in the Toolchain studies privilege composition across MCP tools
~\cite{parasites-mcp}. ETDI signs immutable definitions and preserves approved version/hash state,
while attested admission authenticates the selected server~\cite{etdi,attested-admission}. Tool
Forge carries validation evidence with an execution capsule~\cite{tool-forge}. FCD adds
source-derived effects for the concrete call and formation-scoped successor authority.

Cordon validates staged task effects before commit~\cite{cordon}. FCD covers an earlier boundary:
the call already exists, but its source can retire before execution. The profile bounds successor
effects; the formation record and lifecycle decide whether that successor may execute. The
\emph{Mind the Gap} workshop paper instead studies environment changes between separate calls
~\cite{mind-the-gap}. MCP benchmarks and rug-pull defenses reject malicious or changed definitions
~\cite{mcpsecbench,mcp-rugpull,shieldmcp,agentic-sok}; FCD also covers descriptor-stable handler
changes within one formation-to-execution interval.

Action-policy systems constrain user intent or context-derived authority
~\cite{agent-permissions,attriguard,conseca,agentspec,progent,camel}. ToolEmu, InjecAgent, and
AgentDojo measure unsafe model choices~\cite{toolemu,injecagent,agentdojo}. ChainCaps attenuates
authority across value flows~\cite{chaincaps}. FCD composes with these controls by carrying their
decision to an interpreter authorized at the same formation point.

Isolation and provenance systems mediate components or bind runtime actions
~\cite{isolategpt,ace,saga,hcp,cava,ai-web-plugins}. FCD retains the interpreters authorized to give
that action meaning. Remote attestation and in-toto can strengthen its artifact evidence
~\cite{rats,in-toto}.

\FloatBarrier

\section{Conclusion}
\label{sec:conclusion}

A valid tool call can change security meaning when execution leaves the descriptor--implementation
relation that shaped it. FCD binds three decisions before effect: whether a target preserves the
call's declared in-scope effects, whether formation authorized that target, and whether lifecycle
state makes it eligible. It prefers the exact source and permits a captured successor only after
completed retirement.

Stock releases and deployment paths reproduced the failure. Reviewed profiles reused effect
evidence across releases, and stock-handler traces connected that evidence to admission and effect.
The policy comparison retained existing approval only for the safe pending call. Lifecycle and
replicated-gateway experiments preserved the same decision through execution. Under the
closed-target policy, later rollout decisions may contract pending authority but cannot expand it.

\label{paper:body-end}

\bibliographystyle{IEEEtran}
\bibliography{references/references}

@misc{mcp-tools,
  author = {{Model Context Protocol}},
  title = {Tools},
  year = {2026},
  howpublished = {\url{https://modelcontextprotocol.io/specification/2026-07-28/server/tools}},
  note = {Specification revision 2026-07-28; accessed September 9, 2026}
}

@misc{mcp-basic,
  author = {{Model Context Protocol}},
  title = {Base Protocol},
  year = {2026},
  howpublished = {\url{https://modelcontextprotocol.io/specification/2026-07-28/basic}},
  note = {Specification revision 2026-07-28; accessed September 9, 2026}
}

@misc{mcp-versioning,
  author = {{Model Context Protocol}},
  title = {Versioning and Compatibility},
  year = {2026},
  howpublished = {\url{https://modelcontextprotocol.io/specification/2026-07-28/basic/versioning}},
  note = {Specification revision 2026-07-28; accessed September 9, 2026}
}

@misc{mcp-ts-client,
  author = {{Model Context Protocol}},
  title = {Build Your First Client},
  year = {2026},
  howpublished = {\url{https://ts.sdk.modelcontextprotocol.io/v2/get-started/first-client.html}},
  note = {TypeScript SDK documentation; accessed September 17, 2026}
}

@misc{github-default,
  author = {{GitHub}},
  title = {Default create\_repository to Private When Visibility Omitted},
  year = {2026},
  howpublished = {\url{https://github.com/github/github-mcp-server/pull/2694}},
  note = {Pull request 2694, merged June 15, 2026; accessed September 9, 2026}
}

@misc{github-release-old,
  author = {{GitHub}},
  title = {{GitHub MCP Server 1.3.0}},
  year = {2026},
  howpublished = {\url{https://github.com/github/github-mcp-server/releases/tag/v1.3.0}},
  note = {Official release; accessed September 9, 2026}
}

@misc{github-release-new,
  author = {{GitHub}},
  title = {{GitHub MCP Server 1.4.0}},
  year = {2026},
  howpublished = {\url{https://github.com/github/github-mcp-server/releases/tag/v1.4.0}},
  note = {Official release; accessed September 9, 2026}
}

@misc{azure-default,
  author = {{Microsoft}},
  title = {Disable publicNetworkAccess by Default},
  year = {2026},
  howpublished = {\url{https://github.com/microsoft/mcp/pull/2170}},
  note = {Pull request 2170; accessed September 9, 2026}
}

@misc{azure-release-old,
  author = {{Microsoft}},
  title = {{Azure.Mcp.Server 2.0.0-beta.31}},
  year = {2026},
  howpublished = {\url{https://github.com/microsoft/mcp/releases/tag/Azure.Mcp.Server-2.0.0-beta.31}},
  note = {Official prerelease; accessed September 9, 2026}
}

@misc{azure-release-new,
  author = {{Microsoft}},
  title = {{Azure.Mcp.Server 2.0.0-beta.32}},
  year = {2026},
  howpublished = {\url{https://github.com/microsoft/mcp/releases/tag/Azure.Mcp.Server-2.0.0-beta.32}},
  note = {Official prerelease; accessed September 9, 2026}
}

@misc{protobuf-evolution,
  author = {{Protocol Buffers}},
  title = {Language Guide (proto 3): Updating a Message Type},
  howpublished = {\url{https://protobuf.dev/programming-guides/proto3/\#updating}},
  note = {Accessed September 9, 2026}
}

@misc{aip-180,
  author = {{Google}},
  title = {{AIP-180}: Backwards Compatibility},
  howpublished = {\url{https://google.aip.dev/180}},
  note = {Accessed September 9, 2026}
}

@inproceedings{zhang-upgrade,
  author = {Yongle Zhang and Junwen Yang and Zhuqi Jin and Utsav Sethi and Kirk Rodrigues and Shan Lu and Ding Yuan},
  title = {Understanding and Detecting Software Upgrade Failures in Distributed Systems},
  booktitle = {Proceedings of the ACM SIGOPS 28th Symposium on Operating Systems Principles},
  year = {2021},
  doi = {10.1145/3477132.3483577},
  note = {\url{https://www.cs.purdue.edu/homes/yonglezh/pub/upgrade-sosp21.pdf}}
}

@inproceedings{upfuzz,
  author = {Ke Han and {Sruthi P C} and Yayu Wang and Yaoxu Song and Bishal Basak Papan and Junwen Yang and Pedro Fonseca and Yongle Zhang},
  title = {{UpFuzz}: Detecting Data Format Incompatibility Bugs during Distributed Storage System Upgrade},
  booktitle = {23rd USENIX Symposium on Networked Systems Design and Implementation (NSDI 26)},
  year = {2026},
  pages = {1225--1242},
  publisher = {USENIX Association},
  note = {\url{https://www.usenix.org/conference/nsdi26/presentation/han}}
}

@misc{cwe-toctou,
  author = {{MITRE}},
  title = {{CWE-367}: Time-of-Check Time-of-Use ({TOCTOU}) Race Condition},
  howpublished = {\url{https://cwe.mitre.org/data/definitions/367.html}},
  note = {Accessed September 9, 2026}
}

@techreport{rats,
  author = {Henk Birkholz and Dave Thaler and Michael Richardson and Ned Smith and Wei Pan},
  title = {Remote {ATtestation} procedureS ({RATS}) Architecture},
  institution = {Internet Engineering Task Force},
  type = {RFC},
  number = {9334},
  year = {2023},
  note = {Informational. \url{https://datatracker.ietf.org/doc/html/rfc9334}}
}

@inproceedings{camel,
  author = {Edoardo Debenedetti and Ilia Shumailov and Tianqi Fan and Jamie Hayes and Nicholas Carlini and Daniel Fabian and Christoph Kern and Chongyang Shi and Andreas Terzis and Florian Tram{\`e}r},
  title = {Defeating Prompt Injections by Design},
  booktitle = {2026 IEEE Conference on Secure and Trustworthy Machine Learning (SaTML)},
  year = {2026},
  note = {Official program: \url{https://satml.org/2026/program/}; preprint: \url{https://arxiv.org/abs/2503.18813}}
}

@misc{hcp,
  author = {Ting Liu},
  title = {From Tool Connection to Execution Control: Benchmarking Security Invariants in {MCP}-Style Agent Runtimes},
  year = {2026},
  howpublished = {\url{https://arxiv.org/abs/2606.29073v1}},
  note = {Preprint, arXiv:2606.29073v1}
}

@misc{cava,
  author = {Zexun Wang},
  title = {{CAVA}: Canonical Action Verification and Attestation for Runtime Governance of Agentic {AI} Systems},
  year = {2026},
  howpublished = {\url{https://arxiv.org/abs/2607.13716v1}},
  note = {Preprint, arXiv:2607.13716v1}
}

@inproceedings{schmiedmayer-webapi,
  author = {Paul Schmiedmayer and Andreas Bauer and Bernd Bruegge},
  title = {Reducing the Impact of Breaking Changes to Web Service Clients During Web {API} Evolution},
  booktitle = {2023 IEEE/ACM 10th International Conference on Mobile Software Engineering and Systems (MOBILESoft)},
  year = {2023},
  doi = {10.1109/MOBILSoft59058.2023.00008},
  note = {\url{https://ieeexplore.ieee.org/document/10172926}}
}

@inproceedings{ajmani-upgrades,
  author = {Sameer Ajmani and Barbara Liskov and Liuba Shrira},
  title = {Modular Software Upgrades for Distributed Systems},
  booktitle = {ECOOP 2006 -- Object-Oriented Programming},
  series = {Lecture Notes in Computer Science},
  volume = {4067},
  pages = {452--476},
  year = {2006},
  doi = {10.1007/11785477_26}
}

@inproceedings{morpheus,
  author = {Karla Saur and Joseph Collard and Nate Foster and Arjun Guha and Laurent Vanbever and Michael Hicks},
  title = {Safe and Flexible Controller Upgrades for {SDNs}},
  booktitle = {Proceedings of the Symposium on SDN Research},
  year = {2016},
  pages = {1--12},
  doi = {10.1145/2890955.2890966}
}

@inproceedings{reitblatt-updates,
  author = {Mark Reitblatt and Nate Foster and Jennifer Rexford and Cole Schlesinger and David Walker},
  title = {Abstractions for Network Update},
  booktitle = {Proceedings of the ACM SIGCOMM 2012 Conference},
  year = {2012},
  pages = {323--334},
  doi = {10.1145/2342356.2342427},
  note = {\url{https://conferences.sigcomm.org/sigcomm/2012/paper/sigcomm/p323.pdf}}
}

@inproceedings{in-toto,
  author = {Santiago Torres-Arias and Hammad Afzali and Trishank Karthik Kuppusamy and Reza Curtmola and Justin Cappos},
  title = {in-toto: Providing Farm-to-Table Guarantees for Bits and Bytes},
  booktitle = {28th USENIX Security Symposium (USENIX Security 19)},
  year = {2019},
  pages = {1393--1410},
  publisher = {USENIX Association},
  note = {\url{https://www.usenix.org/conference/usenixsecurity19/presentation/torres-arias}}
}

@inproceedings{toolemu,
  author = {Yangjun Ruan and Honghua Dong and Andrew Wang and Silviu Pitis and Yongchao Zhou and Jimmy Ba and Yann Dubois and Chris J. Maddison and Tatsunori Hashimoto},
  title = {Identifying the Risks of {LM} Agents with an {LM}-Emulated Sandbox},
  booktitle = {International Conference on Learning Representations},
  year = {2024},
  note = {\url{https://proceedings.iclr.cc/paper_files/paper/2024/hash/7274ed909a312d4d869cc328ad1c5f04-Abstract-Conference.html}}
}

@inproceedings{injecagent,
  author = {Qiusi Zhan and Zhixiang Liang and Zifan Ying and Daniel Kang},
  title = {{InjecAgent}: Benchmarking Indirect Prompt Injections in Tool-Integrated Large Language Model Agents},
  booktitle = {Findings of the Association for Computational Linguistics: ACL 2024},
  year = {2024},
  pages = {10471--10506},
  doi = {10.18653/v1/2024.findings-acl.624}
}

@inproceedings{agentdojo,
  author = {Edoardo Debenedetti and Jie Zhang and Mislav Balunovi{\'c} and Luca Beurer-Kellner and Marc Fischer and Florian Tram{\`e}r},
  title = {{AgentDojo}: A Dynamic Environment to Evaluate Prompt Injection Attacks and Defenses for {LLM} Agents},
  booktitle = {Advances in Neural Information Processing Systems 37, Datasets and Benchmarks Track},
  year = {2024},
  doi = {10.52202/079017-2636},
  note = {\url{https://proceedings.neurips.cc/paper_files/paper/2024/hash/97091a5177d8dc64b1da8bf3e1f6fb54-Abstract-Datasets_and_Benchmarks_Track.html}}
}

@inproceedings{attriguard,
  author = {Yu He and Haozhe Zhu and Yiming Li and Shuo Shao and Hongwei Yao and Zhihao Liu and Zhan Qin},
  title = {{AttriGuard}: Defeating Indirect Prompt Injection in {LLM} Agents via Causal Attribution of Tool Invocations},
  booktitle = {35th USENIX Security Symposium (USENIX Security 26)},
  year = {2026},
  pages = {1547--1566},
  publisher = {USENIX Association},
  note = {\url{https://www.usenix.org/conference/usenixsecurity26/presentation/he-yu}}
}

@inproceedings{conseca,
  author = {Lillian Tsai and Eugene Bagdasarian},
  title = {Contextual Agent Security: A Policy for Every Purpose},
  booktitle = {Proceedings of the Workshop on Hot Topics in Operating Systems},
  year = {2025},
  pages = {8--17},
  doi = {10.1145/3713082.3730378},
  note = {\url{https://sigops.org/s/conferences/hotos/2025/papers/hotos25-100.pdf}}
}

@inproceedings{agentspec,
  author = {Haoyu Wang and Christopher M. Poskitt and Jun Sun},
  title = {{AgentSpec}: Customizable Runtime Enforcement for Safe and Reliable {LLM} Agents},
  booktitle = {2026 IEEE/ACM 48th International Conference on Software Engineering (ICSE)},
  year = {2026},
  pages = {2938--2950},
  publisher = {ACM},
  doi = {10.1145/3744916.3764546}
}

@misc{progent,
  author = {Tianneng Shi and Jingxuan He and Zhun Wang and Hongwei Li and Linyu Wu and Wenbo Guo and Dawn Song},
  title = {{Progent}: Securing {AI} Agents with Privilege Control},
  year = {2026},
  howpublished = {\url{https://arxiv.org/abs/2504.11703}},
  note = {Preprint, arXiv:2504.11703}
}

@inproceedings{isolategpt,
  author = {Yuhao Wu and Franziska Roesner and Tadayoshi Kohno and Ning Zhang and Umar Iqbal},
  title = {{IsolateGPT}: An Execution Isolation Architecture for {LLM}-Based Agentic Systems},
  booktitle = {Network and Distributed System Security Symposium},
  year = {2025},
  doi = {10.14722/ndss.2025.241131},
  note = {\url{https://www.ndss-symposium.org/ndss-paper/isolategpt-an-execution-isolation-architecture-for-llm-based-agentic-systems/}}
}

@inproceedings{ace,
  author = {Evan Li and Tushin Mallick and Evan Rose and William Robertson and Alina Oprea and Cristina Nita-Rotaru},
  title = {{ACE}: A Security Architecture for {LLM}-Integrated App Systems},
  booktitle = {Network and Distributed System Security Symposium},
  year = {2026},
  doi = {10.14722/ndss.2026.230352},
  note = {\url{https://www.ndss-symposium.org/ndss-paper/ace-a-security-architecture-for-llm-integrated-app-systems/}}
}

@inproceedings{saga,
  author = {Georgios Syros and Anshuman Suri and Jacob Ginesin and Cristina Nita-Rotaru and Alina Oprea},
  title = {{SAGA}: A Security Architecture for Governing {AI} Agentic Systems},
  booktitle = {Network and Distributed System Security Symposium},
  year = {2026},
  doi = {10.14722/ndss.2026.230869},
  note = {\url{https://www.ndss-symposium.org/ndss-paper/saga-a-security-architecture-for-governing-ai-agentic-systems/}}
}

@inproceedings{agentic-sok,
  author = {Juhee Kim and Wenbo Guo and Dawn Song},
  title = {{SoK}: Attack and Defense Landscape of Agentic {AI} Systems},
  booktitle = {35th USENIX Security Symposium (USENIX Security 26)},
  year = {2026},
  pages = {6047--6066},
  publisher = {USENIX Association},
  note = {\url{https://www.usenix.org/conference/usenixsecurity26/presentation/kim-juhee-agentic}}
}

@misc{mcpsecbench,
  author = {Yixuan Yang and Cuifeng Gao and Daoyuan Wu and Yufan Chen and Yingjiu Li and Shuai Wang},
  title = {{MCPSecBench}: A Systematic Security Benchmark and Playground for Testing Model Context Protocols},
  year = {2025},
  howpublished = {\url{https://arxiv.org/abs/2508.13220}},
  note = {Technical report, arXiv:2508.13220v3}
}

@misc{etdi,
  author = {Manish Bhatt and Vineeth Sai Narajala and Idan Habler},
  title = {{ETDI}: Mitigating Tool Squatting and Rug Pull Attacks in Model Context Protocol ({MCP}) by Using {OAuth}-Enhanced Tool Definitions and Policy-Based Access Control},
  year = {2025},
  howpublished = {\url{https://arxiv.org/abs/2506.01333}},
  note = {Preprint, arXiv:2506.01333}
}

@misc{mind-the-gap,
  author = {Derek Lilienthal and Sanghyun Hong},
  title = {Mind the Gap: Time-of-Check to Time-of-Use Vulnerabilities in {LLM}-Enabled Agents},
  year = {2025},
  howpublished = {NeurIPS 2025 Workshop on Machine Learning for Systems ({MLforSystems})},
  note = {Workshop extended abstract. \url{https://mlforsystems.org/assets/papers/neurips2025/paper25.pdf}}
}

@misc{tool-forge,
  author = {Swanand Rao},
  title = {Tool Forge: A Validation-Carrying Toolchain for Governed Agentic Execution},
  year = {2026},
  howpublished = {\url{https://arxiv.org/abs/2605.28000}},
  note = {Preprint, arXiv:2605.28000}
}

@misc{attested-admission,
  author = {Alfredo Metere},
  title = {Attested Tool-Server Admission: A Security Extension to the Model Context Protocol},
  year = {2026},
  howpublished = {\url{https://arxiv.org/abs/2605.24248}},
  note = {Preprint, arXiv:2605.24248}
}

@misc{mcp-rugpull,
  author = {Saeid Jamshidi and Arghavan Moradi Dakhel and Kawser Wazed Nafi and Foutse Khomh},
  title = {Semantic Attacks on Tool-Augmented {LLMs}: Securing the Model Context Protocol Against Descriptor-Level Manipulation},
  year = {2025},
  howpublished = {\url{https://arxiv.org/abs/2512.06556}},
  note = {Preprint, arXiv:2512.06556v2}
}

@inproceedings{shieldmcp,
  author = {Saurabh Yergattikar},
  title = {Securing the Tool Layer: A Threat Taxonomy and Runtime Defense Framework for Model Context Protocol Deployments},
  booktitle = {Proceedings of the 64th Annual Meeting of the Association for Computational Linguistics: Industry Track},
  year = {2026},
  pages = {865--871},
  doi = {10.18653/v1/2026.acl-industry.58},
  note = {\url{https://aclanthology.org/2026.acl-industry.58/}}
}

@misc{temporal-worker-versioning,
  author = {{Temporal Technologies}},
  title = {Worker Versioning},
  year = {2026},
  howpublished = {\url{https://docs.temporal.io/production-deployment/worker-deployments/worker-versioning}},
  note = {Official documentation; accessed September 14, 2026}
}

@misc{istio-routing,
  author = {{Istio Authors}},
  title = {Request Routing},
  year = {2026},
  howpublished = {\url{https://istio.io/latest/docs/tasks/traffic-management/request-routing/}},
  note = {Official documentation; accessed September 14, 2026}
}

@misc{spiffe-standard,
  author = {{SPIFFE Project}},
  title = {The {SPIFFE} Standard},
  year = {2026},
  howpublished = {\url{https://spiffe.io/docs/latest/spiffe-specs/}},
  note = {Official specifications; accessed September 14, 2026}
}

@inproceedings{chubby,
  author = {Mike Burrows},
  title = {The Chubby Lock Service for Loosely-Coupled Distributed Systems},
  booktitle = {7th USENIX Symposium on Operating Systems Design and Implementation (OSDI 06)},
  year = {2006},
  pages = {335--350},
  publisher = {USENIX Association},
  note = {\url{https://www.usenix.org/conference/osdi-06/presentation/chubby-lock-service-loosely-coupled-distributed-systems}}
}

@article{kitsune,
  author = {Christopher M. Hayden and Karla Saur and Edward K. Smith and Michael Hicks and Jeffrey S. Foster},
  title = {Kitsune: Efficient, General-Purpose Dynamic Software Updating for {C}},
  journal = {ACM Transactions on Programming Languages and Systems},
  year = {2014},
  volume = {36},
  number = {4},
  pages = {13:1--13:38},
  doi = {10.1145/2629460}
}

@misc{postgres-mvcc,
  author = {{PostgreSQL Global Development Group}},
  title = {Concurrency Control: Introduction},
  year = {2026},
  howpublished = {\url{https://www.postgresql.org/docs/18/mvcc-intro.html}},
  note = {PostgreSQL 18 documentation; accessed September 14, 2026}
}

@inproceedings{chaincaps,
  author = {Xiaochong Jiang and Shiqi Yang and Ziwei Li and Lifei Liu and Haoran Yu and Yichen Liu},
  title = {{ChainCaps}: Composition-Safe Tool-Using Agents via Monotonic Capability Attenuation},
  booktitle = {Second Workshop on Agents in the Wild: Safety, Security, and Beyond (AIWILD), ICML 2026},
  year = {2026},
  note = {\url{https://openreview.net/forum?id=KtrAm71ER9}}
}

@misc{contextforge,
  author = {{IBM}},
  title = {{ContextForge}: An {AI} Gateway, Registry, and Proxy for {MCP}, {A2A}, and {REST/gRPC} APIs},
  year = {2026},
  howpublished = {Version 1.0.9; \url{https://github.com/IBM/mcp-context-forge/releases/tag/v1.0.9}},
  note = {Official release; accessed September 15, 2026}
}

@misc{reference-git-option-advisory,
  author = {{Model Context Protocol Authors}},
  title = {Argument Injection in \texttt{git\_diff} and \texttt{git\_checkout} Allows Overwriting Local Files},
  year = {2025},
  howpublished = {GitHub Security Advisory GHSA-9xwc-hfwc-8w59; \url{https://github.com/modelcontextprotocol/servers/security/advisories/GHSA-9xwc-hfwc-8w59}},
  note = {Accessed September 24, 2026}
}

@misc{dbhub-readonly-advisory,
  author = {{Bytebase}},
  title = {Read-Only Mode Does Not Prevent Database Writes},
  year = {2026},
  howpublished = {GitHub Security Advisory GHSA-mwwr-p57h-56pf; \url{https://github.com/bytebase/dbhub/security/advisories/GHSA-mwwr-p57h-56pf}},
  note = {Accessed September 24, 2026}
}

@misc{kubernetes-deployment,
  author = {{Kubernetes Authors}},
  title = {Deployments},
  year = {2026},
  howpublished = {\url{https://kubernetes.io/docs/concepts/workloads/controllers/deployment/}},
  note = {Official documentation; accessed September 15, 2026}
}

@misc{kubernetes-service,
  author = {{Kubernetes Authors}},
  title = {Service},
  year = {2026},
  howpublished = {\url{https://kubernetes.io/docs/concepts/services-networking/service/}},
  note = {Official documentation; accessed September 15, 2026}
}

@inproceedings{parasites-mcp,
  author = {Shuli Zhao and Qinsheng Hou and Zihan Zhan and Yanhao Wang and Yuchong Xie and Yu Guo and Libo Chen and Shenghong Li and Zhi Xue},
  title = {Parasites in the Toolchain: A Large-Scale Analysis of Attacks on the {MCP} Ecosystem},
  booktitle = {2026 IEEE Symposium on Security and Privacy (SP)},
  year = {2026},
  note = {\url{https://arxiv.org/abs/2509.06572}}
}

@inproceedings{agent-permissions,
  author = {Yuhao Wu and Ke Yang and Franziska Roesner and Tadayoshi Kohno and Ning Zhang and Umar Iqbal},
  title = {Towards Automating Data Access Permissions in {AI} Agents},
  booktitle = {2026 IEEE Symposium on Security and Privacy (SP)},
  year = {2026},
  note = {\url{https://homes.cs.washington.edu/~franzi/pdf/wu-agentperms-sp26.pdf}}
}

@inproceedings{rebound,
  author = {Quinn Burke and Anjo Vahldiek-Oberwagner and Michael Swift and Patrick McDaniel},
  title = {It's a Feature, Not a Bug: Secure and Auditable State Rollback for Confidential Cloud Applications},
  booktitle = {2026 IEEE Symposium on Security and Privacy (SP)},
  year = {2026},
  note = {\url{https://arxiv.org/abs/2511.13641}}
}

@inproceedings{racedb,
  author = {An Chen and Yonghwi Kwon and Kyu Hyung Lee},
  title = {{RACEDB}: Detecting Request Race Vulnerabilities in Database-Backed Web Applications},
  booktitle = {2025 IEEE Symposium on Security and Privacy (SP)},
  pages = {939--955},
  year = {2025},
  doi = {10.1109/SP61157.2025.00029},
  note = {\url{https://yonghwi-kwon.github.io/data/racedb_sp25.pdf}}
}

@inproceedings{ai-web-plugins,
  author = {Yigitcan Kaya and Anton Landerer and Stijn Pletinckx and Michelle Zimmermann and Christopher Kruegel and Giovanni Vigna},
  title = {When {AI} Meets the Web: Prompt Injection Risks in Third-Party {AI} Chatbot Plugins},
  booktitle = {2026 IEEE Symposium on Security and Privacy (SP)},
  year = {2026},
  note = {\url{https://arxiv.org/abs/2511.05797}}
}

@inproceedings{symdiff,
  author = {Shuvendu K. Lahiri and Chris Hawblitzel and Ming Kawaguchi and Henrique Reb{\^e}lo},
  title = {{SYMDIFF}: A Language-Agnostic Semantic Diff Tool for Imperative Programs},
  booktitle = {Computer Aided Verification (CAV)},
  series = {Lecture Notes in Computer Science},
  volume = {7358},
  pages = {712--717},
  year = {2012},
  doi = {10.1007/978-3-642-31424-7_54}
}

@inproceedings{idise,
  author = {Neha Rungta and Suzette Person and Joshua Branchaud},
  title = {A Change Impact Analysis to Characterize Evolving Program Behaviors},
  booktitle = {28th IEEE International Conference on Software Maintenance (ICSM)},
  year = {2012},
  doi = {10.1109/ICSM.2012.6405261}
}

@misc{cordon,
  author = {Zheng Chen and Hanqing Liu and Duling Xu and Dong Dong and Jialin Li and Bangzheng Pu and Jidong Zhai},
  title = {Cordon: Semantic Transactions for Tool-Using {LLM} Agents},
  year = {2026},
  howpublished = {arXiv:2606.17573},
  note = {Preprint; \url{https://arxiv.org/abs/2606.17573}}
}

\appendices
\section{Registration Completeness}
\label{app:registration}

This appendix defines the registration premise used by the conditional semantic consequence and
documents its evaluated Kubernetes realization. It separates two questions. FCD determines whether
a pending call may execute on a registered target. The deployment determines which runtime inputs
constitute that target. The latter decision must cover every input that can change the
security-relevant meaning of a call.

\subsection{Authenticated registration manifest}

For each tool epoch, the operator signs a canonical manifest containing the tool identity,
descriptor digest, executable artifact digest, deployment-template generation, configuration
digests, feature-flag snapshot identifier, downstream service identities or API revisions, and
policy sequence. The manifest represents secrets by their immutable version identifiers.
The registry stores the signed manifest and derives the epoch identity from its canonical contents.
Consequently, changing any covered field produces a different epoch across both new-image and
same-image deployments.

The manifest defines the authenticated scope of semantic preservation. Every registered input that
can change authorization, defaults, validation, destination, visibility, or another
security-relevant effect belongs to this scope. The operator declares these inputs, and the signed
manifest makes that declaration auditable. Covered changes force a new epoch at deployment and
admission. Configuration-read tracing and dependency declarations support audits for omitted
dependencies.

The registrar also asserts that the source interpreter implements the registered contract under
these covered inputs. The manifest binds this assertion and its review evidence to the epoch. A
profile-guided certificate additionally binds the analyzed source, profile, generated summary, and
concrete execution context to the same manifest.

These mechanisms make omissions visible to review. The operator defines the complete manifest
schema for the protected tool path. Services with a bounded configuration and dependency surface
can review that schema directly. Dynamic per-request dependencies use the signed snapshots or
enforced contracts described below.

\subsection{Example: Kubernetes deployment}

Our evaluated deployments contain two independently addressed FCD gateway Pods, three etcd
members, a registered backend fixture, and a same-image fixture with a different
security-relevant configuration. The client verifies TLS to each gateway. The local deployment ran
on one physical kind node. The deployment-transfer experiment ran k3s on three distinct EC2 guest
VMs. It placed one etcd member on each VM and the gateways on separate VMs.

The registration manifest binds the stable tool and descriptor to its artifact, configuration
digest, endpoint, and lifecycle state. Formation through gateway A produces an authenticated
capability for that relation. Execution through gateway B verifies the capability and atomically
creates a durable lease marker in etcd. The transaction compares the epoch lifecycle revision,
selected assertion identity, revocation-policy revision, and target identity. Admission racing
with drain or revocation therefore has one serial outcome. Registry unavailability stops the
request before backend dispatch.

The gateway sends the selected artifact and configuration identities to the final fixture path.
The fixture checks them in the request that records the effect. This experimental fence prevents a
Service change from separating identity validation from use. A production realization can replace
it with a backend-native guard or an authenticated sidecar that mediates the only effect path.
Workload identity and remote attestation can authenticate that guard. Fence authority comes from
this authenticated final path; Kubernetes labels and identity endpoints provide observation metadata.

Section~\ref{sec:evaluation} reports the local and EC2 safety outcomes. The audit joins each
admitted or blocked history to its capability and lease record. The EC2 topology records distinct
Kubernetes UIDs, machine identities, private addresses, and provider identities across three guest
VMs in one availability zone.

The initial registry updated one shared epoch record for each acquire and release. Under contention,
that record caused repeated etcd transaction conflicts and range operations. The optimized registry
compares the epoch revision but creates a unique marker per admitted lease. Release deletes that
marker. Retirement first changes the epoch to \texttt{DRAINING} and then scans active markers before
advancing to \texttt{STALE}. A verified-capability cache avoids repeated
signature work while rechecking request context, expiry, route, artifact, configuration, and epoch
state. The optimized safety replication reproduced every safety outcome. At
high concurrency, Figure~\ref{fig:distributed-scaling} compares this design with the shared-record
run.

Production admission can additionally require digest-pinned images and a signed manifest reference.
An admission controller can resolve immutable ConfigMap, Secret-version, and flag-snapshot
identities, recompute their digests, and stamp the epoch into workload identity. This prevents a Pod
whose registered fields differ from the manifest from joining a backend pool. Manifest-schema
review records the mapping from security-relevant configuration reads and dependencies to these
enforced fields.

\subsection{Dynamic flags and dependencies}

Mutable feature flags require immutable, signed snapshots. Formation records the snapshot digest,
and the serving workload proves the same digest at admission or at the backend fence. A flag change
that can alter tool meaning publishes a new snapshot and epoch. Arbitrary per-request flag reads
use an authenticated, versioned snapshot before joining the registered pool.

Downstream services follow the same rule. The manifest binds either an immutable downstream
revision or a contract version enforced at the boundary. Security-relevant downstream defaults
enter the semantic-preservation set through that revision or contract version.

The conditional consequence concerns request interpretation under registered execution inputs. Mutable
application data and external authorization policy compose through their own consistency and
authorization mechanisms.

Operationally, an image, configuration, or covered flag change registers a new epoch. Only a
workload matching its manifest joins the registered pool. A pending call executes its exact
epoch, uses a captured edge only after safe retirement, or rejects before effect. These rules expose
covered registration failures at deployment and admission. The operator defines the manifest
schema. The profile author defines automated transition analysis, while a compatibility issuer
reviews escalations. FCD enforces the resulting formation record. The semantic-preservation set
equals the security-relevant inputs captured by the manifest and the effects modeled by its profile.

\subsection{Reviewed profile acceptance envelopes}
\label{app:profile-envelopes}

Table~\ref{tab:profile-envelopes} records the source structures, argument domains, effect bounds,
and stopping conditions reviewed for the four profiles. These entries define the scope of the
$\mathsf{ActualInScopeEffects}\subseteq S$ premise. Every profile also requires the authenticated
principal and destination, registered artifacts and configuration, declared dependencies, and
final-hop identity described in Section~\ref{sec:security-analysis}.

\begin{table*}[t]
\centering\scriptsize
\setlength{\tabcolsep}{2.5pt}
\caption{Reviewed acceptance envelopes. \texttt{UNKNOWN} produces no certificate. Rejection stops
summary registration.}
\label{tab:profile-envelopes}
\begin{tabularx}{\textwidth}{@{}>{\raggedright\arraybackslash}p{0.105\textwidth}>{\raggedright\arraybackslash}p{0.235\textwidth}>{\raggedright\arraybackslash}p{0.18\textwidth}>{\raggedright\arraybackslash}p{0.19\textwidth}X@{}}
\toprule
Profile & Accepted source structure & Supported arguments & Modeled effects & \texttt{UNKNOWN}, rejection, or escalation \\
\midrule
Reference Git / Python & One \texttt{git\_diff}; direct \texttt{target} flow to one \texttt{repo.git.diff}; optional dash-prefix rejection before the sink & Non-dash targets; exact \texttt{--output=path} family; verified pre-sink dash rejection & Normal diff; filesystem write; handler rejection & Other unpatched options are \texttt{UNKNOWN}; missing or ambiguous sink structure rejects \\
DBHub / JavaScript & Unique classifier and handler functions; every statement reaches the classifier; failed guard returns before \texttt{executeSQL} & Single \texttt{SELECT}; PRAGMA read; exact \texttt{PRAGMA user\_version = integer} family & SQL read; SQLite mutation; handler rejection & Multiple or unsupported statements are \texttt{UNKNOWN}; classifier, keyword, or guard-flow drift rejects or escalates \\
GitHub / Go & One \texttt{CreateRepository}; Boolean \texttt{private} parser; direct value flow through the request to one create sink & Contract-bound name; omitted or explicit Boolean \texttt{private} & Authenticated repository creation; private or public exposure & Parser, default, type, or request-flow mismatch rejects; names outside the contract are \texttt{UNKNOWN} \\
Azure / C\# & Legacy hard-coded enabled shape or complete option wiring to one ARM deployment sink; partial wiring rejects & Contract-bound resource; omission in either shape; explicit Boolean only in the wired shape & Redis creation intent; public network enabled or disabled & Explicit option on the legacy shape is \texttt{UNKNOWN}; partial, duplicate, or missing wiring rejects \\
\bottomrule
\end{tabularx}
\end{table*}

Profile authors review and version these envelopes with the analysis code. The automated
pipeline applies the envelope to each registered release. A release outside the envelope enters
explicit escalation instead of receiving a generated certificate.

\end{document}